\documentclass{article}
\usepackage{iclr2026_conference,times}
\iclrfinalcopy

\usepackage[dvipsnames]{xcolor}
\definecolor{citecolor}{HTML}{1520a6}
\definecolor{linkcolor}{HTML}{900603}
\definecolor{bordo}{HTML}{ff0040}

\usepackage{hyperref}
\hypersetup{
  colorlinks=true,
  citecolor=citecolor,
  linkcolor=linkcolor,
  urlcolor=black,
}

\usepackage{graphicx}
\usepackage{subcaption}
\usepackage{mathtools}
\usepackage{microtype}
\usepackage{float}
\usepackage{url}
\usepackage{booktabs}
\usepackage{enumitem}
\usepackage{amsthm}
\usepackage{amsmath}
\usepackage{amssymb}
\usepackage{tikz}
\usepackage{cleveref}
\usepackage{newtxmath}
\usepackage{wrapfig}
\usepackage{algorithm}
\usepackage[noend]{algpseudocode}
\usepackage{float}
\usepackage{fontawesome5}

\def\ThreeTabRef#1#2#3{Tables \ref{#1}, \ref{#2}, \ref{#3}}

\usepackage{standalone}
\usepackage{tikz}
\usetikzlibrary{positioning}
\usetikzlibrary{arrows}
\usetikzlibrary{calc,fit}
\usetikzlibrary{shapes.geometric}
\usetikzlibrary{shapes.misc}
\usetikzlibrary{decorations.pathmorphing}
\usetikzlibrary{decorations.pathreplacing}

\usepackage{subcaption}

\usepackage{cleveref}
\usepackage{enumitem}

\renewcommand{\eqref}[1]{(\ref{#1})}

\usepackage[createShortEnv,conf={no link to proof, text
link},commandRef=Cref]{proof-at-the-end}
\usepackage{wrapfig}

\makeatother

\makeatletter
\crefname{section}{\S\@gobble}{\S\@gobble}
\crefname{subsection}{\S\@gobble}{\S\@gobble}
\crefname{proposition}{Prop.}{Props.}
\crefname{figure}{Fig.}{Figs.}
\crefname{table}{Table}{Tables}

\def\paragraph{\@startsection{paragraph}{4}{\z@}{0ex}{-1em}{\normalsize\bf}}

\makeatother

\newcommand{\std}[1]{\scriptsize $\pm$#1}

\title{
  Multi-Marginal Flow Matching \\with Adversarially Learnt Interpolants

}

\author{%
  Oskar Kviman\thanks{
    Equal contribution. Correspondence to
    \texttt{k.tamogashev@sms.ed.ac.uk}
  } \\
  KTH \\
  \And
  \hspace{15mm} Kirill Tamogashev\footnotemark[1] \\
  \hspace{15mm} University of Edinburgh \\
  \And
  Nicola Branchini \\
  University of Edinburgh \\
  \AND 
  Víctor Elvira \\
  University of Edinburgh \\
  \And
  \hspace{13mm} Jens Lagergren \\
  \hspace{13mm} KTH \\
  \And
  \hspace{20mm} Esmeralda S. Whitammer\hspace{-20mm} \\
  \hspace{20mm} University of Edinburgh \\
  \hspace{20mm} CIFAR Fellow
  \AND
  \hspace{17mm} \href{https://github.com/mmacosha/adversarially-learned-interpolants}{\textcolor{citecolor}{\faGithub \; \texttt{mmacosha/adversarially-learned-interpolants}}}
}

\begin{document}

\maketitle

\begin{abstract}
  \looseness=-1
  Learning the dynamics of a process given sampled observations at
  several time points is an important but difficult task in many
  scientific applications.
  When no ground-truth trajectories are available, but one has only
  snapshots of data taken at discrete time steps, the problem of
  modelling the dynamics, and thus inferring the underlying
  trajectories, can be solved by multi-marginal generalisations of
  flow matching algorithms.
  This paper introduces a novel flow matching method that overcomes
  the limitations of existing multi-marginal trajectory inference algorithms.
  Our proposed method, ALI-CFM, uses a GAN-inspired adversarial loss
  to fit neurally parameterised interpolant curves between source and
  target points such that the marginal distributions at intermediate
  time points are close to the observed distributions.
  The resulting interpolants are smooth trajectories that, as we
  show, are unique under mild assumptions.
  These interpolants are subsequently marginalised by a flow matching
  algorithm, yielding a trained vector field for the underlying dynamics.
  ALI-CFM outperforms existing baselines on spatial transcriptomics and
  cell tracking problems, while performing on par with them on
  single-cell trajectory prediction, which showcases its versatility and scalability.
\end{abstract}

\section{Introduction}
\looseness=-1
Modelling the time-dependent dynamics of a system given experimental
observations is a central task in many scientific problems in biology (see
\cite{schiebinger2019optimal, bunne2023learning}), medicine (see
\cite{oeppen2002broken, hay2021estimating}), and other areas. The problem
involves a collection of data snapshots taken  
\begin{wrapfigure}[13]{r}{0.41\textwidth}
  \vspace*{-1em}
  \centering
  \begin{tikzpicture}[>=stealth]

\draw[->] (-0.,0) -- (8,0) ;

\node[above] at (0.5,0) {$t=0$};
\node[above] at (4, -0.1) {$t=t_j$};
\node[above] at (8 - 0.5,0) {$t=1$};

\node[below, blue] at (0.5 - 0.7,-0.15) {$q_{0}$};
\node[below, blue] at (3.5 - 0.7,-.15) {$q_{t_j}$};
\node[below, blue] at (7.5 - 0.7,-0.15) {$q_1$};

\node[below,blue] at (0.5,2.5) {$x_{0}$};
\node[below,blue] at (4,2.5) {$x_{t_j}$};
\node[below,blue] at (7.5,2.5) {$x_1$};

\draw[domain=-1:1,smooth,variable=\x,blue,thick,dotted] 
  plot ({0.5+\x},{0.8*exp(-(\x*\x)/0.1) - 1.25});
\draw[domain=-1:1,smooth,variable=\x,red,thick, dashed] 
  plot ({0.5+\x},{0.8*exp(-(\x*\x)/0.1) - 1.25});

\draw[domain=2.5:5.5,smooth,variable=\x,blue,thick,dotted]
  plot ({\x},{0.8*exp(-((\x-3.5)^2)/0.1) + 0.8*exp(-((\x-4.5)^2)/0.1) - 1.25});
\draw[domain=2.5:5.5,smooth,variable=\x,red,thick,dashed]
  plot ({\x+0.2},{0.8*exp(-((\x-3.4)^2)/0.15) + 0.8*exp(-((\x-4.5)^2)/0.1) - 1.25});
  
\draw[domain=-1:1,smooth,variable=\x,blue,thick,dotted] 
  plot ({7.5+\x},{0.8*exp(-(\x*\x)/0.1) - 1.25});
\draw[domain=-1:1,smooth,variable=\x,red,thick,dashed] 
  plot ({7.5+\x},{0.8*exp(-(\x*\x)/0.1) - 1.25});

\draw[red!70!red,dashed] (0.75,0.75) -- (3.75,0.6);
\draw[red!70!red,dashed] (0.25,1.) -- (4.25,0.9);
\draw[red!70!red,dashed] (0.75,1.5) -- (4,1.25);
\draw[red!70!red,dashed] (0.25,1.75) -- (3.75,1.5);

\draw[red!70!red,dashed] (3.75,0.6) -- (7.25,0.75);
\draw[red!70!red,dashed] (7.75,1.) -- (4.25,0.9);
\draw[red!70!red,dashed] (7.25,1.5) -- (4,1.25);
\draw[red!70!red,dashed] (7.75,1.75) -- (3.75,1.5)node[midway,above] {$G_\phi(x_0, x_1, t)$};

\node[circle,fill=blue!70!blue,inner sep=1pt] at (0.75,0.75) {};
\node[circle,fill=blue!70!blue,inner sep=1pt] at (0.25,1) {};
\node[circle,fill=blue!70!blue,inner sep=1pt] at (0.75,1.5) {};
\node[circle,fill=blue!70!blue,inner sep=1pt] at (0.25,1.75) {};

\node[circle,fill=blue!70!blue,inner sep=1pt] at (4.25,0.55) {};
\node[circle,fill=blue!70!blue,inner sep=1pt] at (3.75,0.8) {};
\node[circle,fill=blue!70!blue,inner sep=1pt] at (4.25,1.3) {};
\node[circle,fill=blue!70!blue,inner sep=1pt] at (3.75,1.55) {};

\node[circle,fill=blue!70!blue,inner sep=1pt] at (7.25,0.75) {};
\node[circle,fill=blue!70!blue,inner sep=1pt] at (7.75,1) {};
\node[circle,fill=blue!70!blue,inner sep=1pt] at (7.25,1.5) {};
\node[circle,fill=blue!70!blue,inner sep=1pt] at (7.75,1.75) {};



\end{tikzpicture}\\[-0.2em]
  \caption{\looseness=-1
    Adversarially learnt interpolants (red curves) follow
    pushforward distributions (red densities) that approximate the
    intermediate-time marginal distributions $q_{t_j}$ (blue) and, by
    construction, have the correct end-marginals $q_0$ and $q_1$.
  }
  \label{fig:ali}
\end{wrapfigure}
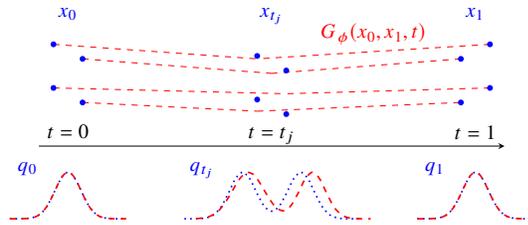
at various time steps that
together provide an empirical account of some process.
Examples of such
processes include recordings of health measurements, evolution of a disease
\citep{waddington1942epigenotype, hay2021estimating}, and time-series
single-cell RNA sequencing data (scRNA-seq; \cite{macosko2015highly,
klein2015droplet}).

Formally, a (deterministic)
system in $\mathbb{R}^n$ can be described by an ordinary differential
equation (ODE) $dx_t=v_t(x_t)\,dt$, where $v_t$ is a time-dependent
vector field. In a bimarginal case where samples are given from a
pair of marginal distributions $q_0,q_1$, we aim to find $v_t$ such
that the ODE's integration map from $t=0$ to $t=1$ pushes $q_0$ to $q_1$.
In the multi-marginal case, the marginal distributions $p_t$ induced
by the dynamics with initial conditions $p_0=q_0$ should also satisfy
intermediate conditions, namely, $p_{t_i} = q_{t_i}$ for a set of
times $0=t_1<t_2<\dots<t_K= 1$. The intermediate distributions
$q_{t_i}$ are provided by datasets of samples.

While this problem can be tackled by applying flow matching
\citep{lipman2023flow, albergo2023stochastic, liu2023flow,
tong2023improving} to learn flows for every pair of consecutive
marginals $q_{t_i},q_{t_{i+1}}$ \citep{tong2023improving}, we show
that such approaches can lead to non-smooth interpolation curves.
Recently, several methods specialising in multi-marginal problems
have been proposed.
\cite{rohbeck2025modeling} proposes to use cubic splines to build
trajectories that pass through samples from intermediate distributions.
However, spline interpolation methods do not seem to scale well to
high dimensions \citep{lee2025multi}.
Another line of work \citep{neklyudov2023computational,
kapusniak2024metric} assumes that the interpolants in each interval
$[t_i,t_{i+1}]$  follow a certain geometry that can be
learnt. These methods fit an interpolant and use it for learning the
vector field of the underlying dynamics.
However, the interpolants are still piecewise and are difficult to marginalise with precision, as we show in \Cref{sec:knot_exp} and \Cref{sec:cell_tracking}
(see, e.g., \Cref{fig:figure1}).

To overcome the limitations of previous methods, we propose to learn
the interpolations between distributions using a GAN-like adversarial
objective \citep{goodfellow2014generative, huanggan2024ganisdead} to
obtain what we call adversarially learnt interpolants (ALIs).
ALI directly matches the target intermediate marginals with those of
the learnt interpolants (\Cref{fig:ali}). The interpolants can then
be marginalised by a conditional flow matching (CFM) method, allowing
us to model complex time-dependent behaviour while using available
intermediate-time information. The full algorithm, called ALI-CFM,
approximates the data distribution at each time step, as opposed to
prior methods that force interpolants to pass through samples
explicitly, which makes our method especially useful in cases where
the provided dataset is noisy.

Our core contributions in this paper are the following:
\begin{enumerate}[left=0pt,nosep,label=(\arabic*)]
  \item We propose a novel technique for learning interpolants using a GAN-like
    adversarial objective. We show that the introduced algorithm can be easily
    used for multi-marginal problems with snapshots in either
    discrete or continuous
    time. We show the versatility of ALI-CFM on both synthetic
    and real-data examples.
  \item We successfully apply our method to a tumour coordinate
    inference problem using
    spatial transcriptomics (ST) data, where our method significantly
    outperforms
    all the existing baselines.
  \item We demonstrate the scalability of ALI-CFM on a scRNA-seq
    trajectory inference problem.
\end{enumerate}

\begin{figure*}[t!]
  \centering
  \includestandalone[width=1\linewidth]{figs_fpi/figure1}
  \caption{
    Comparison of CFM \citep{tong2023improving}, MFM
    \citep{kapusniak2024metric},
    MMFM \citep{rohbeck2025modeling} and our ALI-CFM method on a
    synthetic 2D `knot' distribution.
    See \Cref{sec:experiments} for details.
  }
  \label{fig:figure1}
\end{figure*}

\section{Method}
\label{sec:method}
\paragraph{Background.}
We summarise CFM, roughly following the setting and exposition of
\cite{lipman2023flow,tong2023improving, pooladian2023multisample}.
All statements below hold under regularity
conditions whose details are not relevant in this paper.

We assume a dynamical system in $\mathbb{R}^n$ given by an ordinary
differential equation (ODE) $dx_t=v_t(x_t)\,dt$. Its integration map $\psi_t:
\mathbb{R}^{n} \rightarrow \mathbb{R}^{n}$ from time 0 to time $t$ satisfies
\begin{equation}
  \frac{d}{dt}\psi_t(x_0) = v_t(\psi_t(x_0)) , \quad \psi_0(x_0) = x_0.
  \label{eq:integration_map}
\end{equation}
The integration map, together with stochastic initial conditions $x_0
\sim p_0$, defines a probability path $p_t=(\psi_t)_\#p_0$, where
$p_t$ is the marginal distribution of $x_t$.

In the bimarginal FM setting, one observes samples from two marginal
distributions, $x_0 \sim q_0$ and $x_1 \sim q_1$, and fixes a vector field
$v_t$ such that its integration map $\psi_t$ satisfies $(\psi_1)_\#q_0=q_1$.
(This vector field is not tractably computable, but is described by
interpolants, as we discuss below.) Having fixed $v_t$ as a target, the goal is
to approximate it with a neural net $u^\theta_t$ with weights $\theta$. This
could be done via the FM objective
\begin{equation}
  L_\text{FM} = \mathbb{E}_{t\sim U[0, 1],x\sim p_t}\| u^\theta_t(x)
  - v_t(x) \|^2_2.
  \label{eq:fm}
\end{equation}
However, since $v_t$ and $p_t$ are intractable to compute and sample
from, the loss \eqref{eq:fm} is intractable.

Instead, one assumes a family of \emph{interpolant} curves, one for every pair
$x_0,x_1$ in the support of $q_0\otimes q_1$. These curves are denoted
$G(x_0,x_1,t)$ and should satisfy $G(x_0,x_1,t)=x_t$ for $t=0,1$. For example,
a linear interpolant is described by $G(x_0,x_1,t)=tx_1+(1-t)x_0$. We define
$v_t(x_t\mid x_0,x_1):=\frac{d}{dt}G(x_0,x_1,t)$. For any joint distribution
$\pi$ over $\mathbb{R}^n\times\mathbb{R}^n$ whose marginals are $q_0$ and
$q_1$, respectively, it can then be shown that the marginal vector field
\begin{equation}
  v_t(x_t) := \mathbb{E}\left[v_t(x_t\mid x_0,x_1)\mid
  x_t=G(x_0,x_1,t)\right], \quad (x_0,x_1)\sim\pi,
  \label{eq:marginal_flow}
\end{equation}
pushes $q_0$ to $q_1$ and can thus be used as the learning target for
$u^\theta_t$. The marginals $p_t$ of the resulting dynamics are
tractably sampled by drawing $(x_0,x_1)\sim\pi$ and setting
$x_t=G(x_0,x_1,t)$. While $v_t$ itself is still not tractable, one
can replace \eqref{eq:fm} by the following \emph{conditional flow
matching} (CFM) objective:
\begin{equation}
  L_\text{CFM} = \mathbb{E}_{t\sim U[0,1],(x_0,x_1)\sim\pi}\|
  u^\theta_t(x_t) - v_t(x_t|x_0,x_1)
  \|^2_2, \text{where $x_t=G(x_0,x_1,t)$}.
  \label{eq:cfm}
\end{equation}
It is easy to show that the gradients of \eqref{eq:cfm} and
\eqref{eq:fm} coincide, and \eqref{eq:cfm} thus provides a tractable
way to learn the target vector field \citep{lipman2023flow}.

The above setting leaves two choices open: the \textbf{coupling}
$\pi$ and the \textbf{interpolants} $G$.
Past work has proposed an independent coupling $\pi=q_0\otimes q_1$,
giving objectives equivalent to those in
\cite{liu2023flow,albergo2023stochastic}, or couplings computed by
(possibly minibatch or entropic) optimal
transport \citep{tong2023improving,pooladian2023multisample}. The
latter has been shown to result in straighter
integration curves and solve the dynamic optimal transport problem.
For the interpolants, linear and trigonometric
\citep{albergo2023stochastic} curves have been proposed, as well as
those trained to pass through areas of high data
distribution density \citep{kapusniak2024metric}. In the
multi-marginal setting, piecewise linear \citep{tong2023improving}
and cubic spline \citep{rohbeck2025modeling} interpolants have been
used. See \Cref{sec:app:related_work} for a discussion.

\subsection{Adversarial learning of interpolants}
We move from the bimarginal to the multi-marginal setting, where data has been
collected from a sequence of $K$ marginal distributions, $x_{t_i}\sim q_{t_i}$
with corresponding time stamps $0 = t_1 < t_2 < \dots < t_K = 1$. Given a
coupling $\pi$ between $q_0$ and $q_1$ and a neural network $f_\phi$, we
consider neural interpolants
\citep{neklyudov2023computational,kapusniak2024metric}
\begin{equation}
  \label{eq:ALIs}
  G_\phi(x_0, x_1, t) = (1-t) x_0 + tx_1 + t(1-t) f_\phi(x_0, x_1, t).
\end{equation}
Our aim is to match the intermediate distributions at $t_i$ of the
interpolants when $(x_0,x_1)\sim\pi$ to the given marginals
$q_{t_i}$, that is, to enforce
\begin{equation}\label{eq:match}
  (G_\phi(\cdot,\cdot,t_i))_\#\pi=q_{t_i}.
\end{equation}
(The parametrisation \eqref{eq:ALIs} guarantees
  $G_\phi(x_0,x_1,t)=x_t$ for $t=0,1$, so \eqref{eq:match} holds
automatically for $i\in\{1,K\}$.)
In order to approximately enforce \eqref{eq:match}, we use an
adversarial learning scheme. Let $D_\gamma(x_t, t)$ be a second
neural network, tasked to discriminate between marginal samples, $x_t
\sim q_t$, and the learnable interpolants in \eqref{eq:ALIs}.
Optimising the min-max GAN objective \citep{goodfellow2014generative}
for each $t_i$,
\begin{equation}
  \min_{G_\phi}\max_{D_\gamma}\underbracket{\mathbb{E}_{(x_0,x_1)
    \sim \pi}\left[\log(1 - D_\gamma(G_\phi(x_0, x_1, t_i),
    t_i))\right] + \mathbb{E}_{q_{t_i}}\left[\log D_\gamma(x_{t_i},
  t_i)\right]}_{L_\text{GAN}(G_\phi, D_\gamma; t_i)},
  \label{eq:minmax}
\end{equation}
is then equivalent, under the assumption of an optimal discriminator, to minimising the Jensen-Shannon divergence
\citep{goodfellow2014generative} between $q_{t_i}$ and, in this case,
$G_\phi(\cdot, \cdot, t_i)_\#\pi$. Notably, our `generators' -- the
interpolants -- are conditioned on scalar-valued time inputs,
associated with the targeted $q_{t_i}$. While the noise in GANs
typically comes from a fixed distribution, in \eqref{eq:minmax} the
pair $(x_0,x_1)\sim\pi$ plays the role of `noise'.

The solutions to the min-max problem in \eqref{eq:minmax} are not unique and
can induce arbitrarily curved interpolants. To this end, we propose
regularising terms, $L_\text{reg}(G_\phi; t)$, in the learning objective that
guarantee unique interpolants. See \Cref{sec:regularisers} for explicit
formulations of the proposed regularisers. In \Cref{alg:ali} we provide
pseudocode for training ALIs.
\begin{algorithm}[H]
\caption{Training adversarially learnt interpolants}
\label{alg:ali}
\begin{algorithmic}[1]
\Require coupling $\pi$, trainable correction network $f_{\phi}$, trainable discriminator $D_\gamma$, regulariser $L_\text{reg}$, regularising weight $\lambda$
\While{Training}
  \State Draw $(x_0,x_1)\sim \pi$, $i\sim \text{Unif}(\{2,...,K-1\})$, $x_{t_i}\sim q_{t_i}$
  \State $G_{\phi}(x_0,x_1,t_i) \gets (1-t_i)\,x_0 + t_i\,x_1 + t_i(1-t_i)\,f_{\phi}(x_0,x_1, t_i)$ \Comment{ \Cref{eq:ALIs}}
  \State $\widehat{L}_\text{GAN} \gets$ estimate GAN loss with $x_{t_i}$ and $G_{\phi}(x_0,x_1,t_i)$ \Comment{ \Cref{eq:minmax}}
  \State $\widehat{L}_\text{reg}\gets$ estimate regularising term \Comment{Regularisers are suggested in \Cref{sec:regularisers}}
  \State $\widehat{L}_\text{ALI} \gets \widehat{L}_\text{GAN} + \lambda \widehat{L}_\text{reg}$ \Comment{\Cref{eq:ali_objective}}
  \State Update $\phi$ and $\gamma$ using gradient $\nabla_{\phi,\gamma}\,\widehat{L}_\text{ALI}$
\EndWhile
\State \Return adversarially learnt interpolants $G_\phi$
\end{algorithmic}
\end{algorithm}

\subsection{Complete method: ALI-CFM} The loss \eqref{eq:minmax} and
the above result motivate the full
\emph{adversarially learnt interpolants (ALI) objective}:
\begin{equation}\label{eq:ali_objective}
  L_\text{ALI}(G_\phi,D_\gamma) =
  \mathbb{E}_{i\sim\text{Unif}(\{2,\dots,K-1\})}\left[
    L_\text{GAN}(G_\phi,D_\gamma;t_i)+\lambda L_\text{reg}(G_\phi;t_i)
  \right],
\end{equation}
where $\lambda>0$ is a regularisation weight.

Once the interpolants $G_\phi$ have been trained using ALI, they can be
marginalised using the CFM objective \eqref{eq:cfm}, using the same coupling
$\pi$.\footnote{The objective requires differentiating the learnt interpolants
  with respect to $t$, which is done with little overhead using autograd. Note
that the interpolants' parameters $\phi$ are fixed at this stage.} At
convergence, this yields a dynamical system, defined by a vector field
$u^\theta_t$, whose marginals $p_t$ at each $t$ match the interpolants'
marginals $(G_\phi(\cdot,\cdot,t))_\#\pi$. If further \eqref{eq:match} has been
satisfied, then the resulting flow solves the multi-marginal transport problem:
$p_{t_i}=q_{t_i}$ for all $i$.

We refer to the complete method as ALI-CFM: it consists of (i) learning
interpolants $G_\phi$ using ALI (\Cref{alg:ali}), then (ii) marginalising them
to yield the time-dependent vector field $u^\theta_t$ (\Cref{alg:ali_cfm}). The
prefixes I- and OT- in the algorithm names specify if the coupling $\pi$ used
is independent or a (minibatch) optimal transport plan, respectively.

\begin{algorithm}[H]
\caption{Training ALI-CFM}
\label{alg:ali_cfm}
\begin{algorithmic}[1]
\Require coupling $\pi$, trained ALI $G_{\phi}$, trainable CFM net $u_t^\theta$
\While{Training}
  \State $(x_0,x_1)\sim \pi$, $t\sim U[0,1]$
  \State $G_{\phi}(x_0,x_1,t) \gets (1-t)\,x_0 + t\,x_1 + t(1-t)\,f_{\phi}(x_0,x_1, t)$ \Comment{ \Cref{eq:ALIs}}
  \State $\frac{d}{dt}{G}_{\phi}(x_0,x_1, t) \gets x_1 - x_0 + t(1-t)\,\frac{d}{dt}{f}_{\phi}(x_0,x_1, t) + (1-2t)\,{f}_{\phi}(x_0,x_1, t)$
  \State $\widehat{L}_\text{ALI-CFM} \gets \|u^\theta_t({G}_{\phi}(x_0,x_1, t)) - \frac{d}{dt}{G}_{\phi}(x_0,x_1, t)\|^2$  \Comment{Estimate CFM loss}
  \State Update $\theta$ using gradient $\nabla_{\theta}\,\widehat{L}_\text{ALI-CFM}$
\EndWhile
\State \Return learned vector field $u_t^\theta$
\end{algorithmic}
\end{algorithm}

\subsection{Regularisers for interpolants}
\label{sec:regularisers}
In this section we propose three regularising terms to use in
\eqref{eq:ali_objective} with provable uniqueness guarantees for the
first two options. See \Cref{sec:scrna_traj} for an ablation study of
the different regularisers.
\paragraph{Linear reference regulariser.}
Here we introduce a regularising term in the ALI learning objective which
penalises deviations from the linear interpolant between coupled samples from
the end-marginals. Letting $\ell(x_0, x_1, t) = (1-t)x_0 + tx_1$, we define
\begin{align}\label{eq:reg}
  L_\text{reg}(G_\phi; t_i) & = \mathbb{E}_{(x_0,x_1) \sim
  \pi}\left[\Vert G_\phi(x_0, x_1, t_i) - \ell(x_0, x_1, t)  \Vert^2 \right].
\end{align}
The problem of matching marginals \eqref{eq:match} while minimising
the regulariser \eqref{eq:reg} enjoys unique solutions:
\begin{theoremE}[][end,restate]
  \label{thm:linear}
  Fix $t\in(0,1)$, $q_t$, and a coupling $\pi$ between $q_0$ and
  $q_1$ such that the distribution  $\ell(\cdot, \cdot, t)_\#\pi$ is
  absolutely continuous (a.c.) w.r.t.\ the Lebesgue measure.
  Then the interpolant
  $G(\cdot,\cdot,t):\mathbb{R}^d\times\mathbb{R}^d\to\mathbb{R}^d$ minimising
  \begin{equation}\label{eq:cost_app}
    \mathbb{E}_{(x_0,x_1) \sim\pi}\|G(x_0,x_1,t)-\ell(x_0, x_1, t)\|^2
  \end{equation}
  subject to $G(\cdot,\cdot,t)_\#\pi=q_{t}$, exists and is unique on
  the support of $\pi$ up to almost-everywhere equality.
\end{theoremE}
\begin{proofE}
  Let $G(\cdot,\cdot,t)$ be any function satisfying the constraint.
  Consider the following joint distribution on $\mathcal{X} \times \mathcal{X}$,

  \begin{equation}
    \Pi(dx_t, dx_{t}^\prime) =(\ell(\cdot,\cdot,t),G(\cdot,\cdot,t))_\#\pi.
  \end{equation}

  The marginals of $\Pi$ over the first and second components are the a.c.
  distributions $\ell(\cdot,\cdot,t)_\#\pi$ and $q_t$, respectively. The
  distribution $\Pi$ is a Kantorovich plan between these two
  marginals whose cost
  is given by \eqref{eq:cost_app}. This cost has a unique minimiser over all
  transport maps $\Pi$ (up to a.e.\ equality on the supports) because the first
  marginal is a.c., and the minimiser has $\Pi$ deterministic over the second
  component given the first, i.e., $\Pi$ is given by $({\rm
  Id},T)_\#\ell(\cdot,\cdot,t)$
  for some function $T: \mathcal{X} \rightarrow \mathcal{X}$.

  This minimum is indeed uniquely achieved by $G=T\circ\ell-\ell$. Conversely,
  the optimal transport plan from $\ell(\cdot,\cdot,t)_\#\pi$ to $q_t$ yields an
  interpolant $G$ in the obvious manner, showing existence.
\end{proofE}
We provide the proof in \Cref{app:proof}. Note that the assumption of
an a.c.\ target interpolant is satisfied under a number of
conditions, e.g., if $\pi$ is the solution to entropic OT with
squared-euclidean cost (or either of its limiting cases:
(nonentropic) OT or the independent coupling) and either $q_0$ or $q_1$ is a.c.

\paragraph{Piecewise linear reference regulariser.}
In settings where the supports of the intermediate marginal distributions
differ from those of the end marginals, the linear reference from the previous
section may restrict $G_\phi$ from accurate modelling of the target
distributions, especially in high dimensions. Also, when $K$ is small, the
learning of $f_\phi$ may benefit from objectives evaluated on the pseudo-time
unit interval, as opposed to on a discrete set of time stamps. As such, we
propose a second regulariser that regresses $G_\phi$ to a piecewise linear
interpolant, marginalised over $t\in[0,1]$
\begin{equation}
  \label{eq:piecewise_reg}
  L_\text{reg}(\phi; t_i) = \mathbb{E}_{t\sim
  U[0,1]}\mathbb{E}_{(x_1, x_{t_i}, x_0) \sim
  \pi_{t_i}}\left[\|G_\phi(x_0, x_1,t) - \ell(x_t|x_0, x_1, x_{t_i},
  t)\|^2\right],
\end{equation}
where we let $\pi_{t_i}$ be a Markov-chained OT coupling
\citep{tong2023improving}
$\pi_{t_i}=\pi(x_1|x_{t_i})\pi(x_{t_i}|x_0)q_0(x_0)$, with
\begin{equation}
  \label{eq:piecewise_interpolant}
  \ell(x_t|x_0, x_1, x_{t_i}, t) =
  \begin{cases}
    \frac{tx_{t_i} + (t_k - t)x_0}{t_i}, \quad t\leq t_i, \\
    \frac{tx_1 + (1 - t)x_{t_i}}{1 - t_i}, \quad t > t_i,
  \end{cases}
\end{equation}
and which comes with the following uniqueness guarantee.
\begin{theoremE}[][end,restate]
  \label{thm:piecewise_linear}
  Fix $t\in(0,1)$, $q_t$, and a Markov–chained OT coupling
  $\pi_{t_i}=\pi(x_1\mid x_{t_i})\,\pi(x_{t_i}\mid x_0)\,q_0(x_0)$
  between $q_0$ and $q_1$. Let $\ell(x_t\mid x_0,x_1,x_{t_i},t)$ be the
  piecewise linear reference map in \eqref{eq:piecewise_interpolant},
  and assume that the distribution
  $\ell(\,\cdot\mid\cdot,\cdot,\cdot,t)_\#{\pi_{t_i}}$ is a.c.
  with respect to the Lebesgue measure. Then the interpolant
  $G(\cdot,\cdot,t):\mathbb{R}^d\times\mathbb{R}^d\to\mathbb{R}^d$ minimising
  \begin{equation}
    \mathbb{E}_{(x_0,x_{t_i},x_1)\sim\pi_{t_i}}
    \big\| G(x_0,x_1,t)-\ell(x_t\mid x_0,x_1,x_{t_i},t)\big\|^2
    \label{eq:ref-pointwise}
  \end{equation}
  subject to $G(\cdot,\cdot,t)_\#{\pi}=q_t$, where $\pi$ is the
  $(x_0,x_1)$-marginal of $\pi_{t_i}$, exists and is unique on the support of
  $\pi$ up to almost-everywhere equality.
  Moreover, if $t$ is averaged according to $U[0,1]$ as in
  \eqref{eq:piecewise_reg},
  \begin{equation}
    L_{\mathrm{reg}}(G; t_i)
    =\mathbb{E}_{t\sim U[0,1]}\,\mathbb{E}_{(x_0,x_{t_i},x_1)\sim\pi_{t_i}}
    \big\| G(x_0,x_1,t)-\ell(x_t\mid x_0,x_1,x_{t_i},t)\big\|^2,
  \end{equation}
  the same $G(\cdot,\cdot,t)$ minimises $L_{\mathrm{reg}}$ for $U$-a.e.\ $t$.
\end{theoremE}
\begin{proofE}
  The proof of the first part of the theorem follows similar
  arguments as in \Cref{thm:linear}. Let $G(\cdot,\cdot,t)$ be any
  function satisfying the constraint.
  Consider the following joint distribution on $\mathcal{X}\times\mathcal{X}$:
  \begin{equation}
    \Pi(dx_t,dx_t') \;=\;
    \big(\,\ell(\cdot\mid\cdot,\cdot,\cdot,t),\;
    G(\cdot,\cdot,t)\,\big)_\#{\pi_{t_i}}.
  \end{equation}

  The marginals of $\Pi$ over the first and second components are the a.c.
  distributions $\ell(\,\cdot\mid\cdot,\cdot,\cdot,t)_\#{\pi_{t_i}}$ and $q_t$,
  respectively. As in \Cref{thm:linear}, the distribution $\Pi$ is a Kantorovich
  plan between these two marginals whose cost is
  \begin{equation}
    \int \|x_t'-x_t\|^2\,d\Pi(x_t,x_t')
    \;=\;
    \mathbb{E}_{(x_0,x_{t_i},x_1)\sim\pi_{t_i}}
    \big\| G(x_0,x_1,t)-\ell(x_t\mid x_0,x_1,x_{t_i},t)\big\|^2,
  \end{equation}
  i.e., the regulariser in \eqref{eq:piecewise_reg}. This cost has a unique
  minimiser over all transport maps $\Pi$ (up to a.e.\ equality on the supports)
  because the first marginal is a.c., and the minimiser has $\Pi$ deterministic
  over the second component given the first, i.e., $\Pi=(\mathrm{Id},T)_\#\mu_t$
  for some function $T:\mathcal{X}\to\mathcal{X}$, where
  $\mu_t=\ell(\cdot\mid\cdot,\cdot,\cdot,t)_\#{\pi_{t_i}}$. This minimum is
  indeed uniquely achieved by
  \begin{equation}
    G(x_0,x_1,t)= T\circ\ell(x_t\mid x_0,x_1,x_{t_i},t),
    \quad (x_0,x_{t_i},x_1) \sim \pi_{t_i},
  \end{equation}
  which proves existence and uniqueness on the support of $\pi$. Since the
  argument is pointwise in $t$, integrating with respect to any $U$ on $[0,1]$
  (as in \eqref{eq:piecewise_reg}) preserves optimality for $U$-a.e.\ $t$.
\end{proofE}
See \Cref{app:proof} for a proof.

\paragraph{Norm of the second derivative as a regulariser.}
An alternative way to introduce regularisation is to add a component that would
enforce a smoothness constraint on the entire interpolant. One way to do that
is to use the integral of the norm of the second derivative of the interpolant.
A similar idea motivated the use of cubic splines in
\citet{rohbeck2025modeling}. In contrast, we directly incorporate the integral
of the norm of the second derivative as a regularisation term:
\begin{equation}\label{eq:2nd_derivative}
  L_\text{reg}(\phi) = \mathbb{E}_{(x_0, x_1) \sim \pi}
  \left[
    \int_0^1 \left\|\frac{\partial^2}{\partial t^2} G_{\phi}(x_0,
    x_1, t)\right\|^2_2 dt.
  \right]
\end{equation}
Computing the second derivative of the neural network exactly would
be prohibitively costly. Instead, we use a numerical approximation of
the second derivative:
\begin{equation}
  \frac{\partial^2}{\partial t^2} G_{\phi}(x_0, x_1, t) \approx
  \frac{ G_{\phi}(x_0, x_1, t + h) + G_{\phi}(x_0, x_1, t - h) - 2
  \cdot G_{\phi}(x_0, x_1, t)}{h^2}.
\end{equation}
We found a Monte Carlo estimate of the integral using three samples
of $t\sim U[0,1]$ to be sufficient.

\section{Related works}
\label{sec:app:related_work}
Here, we describe previously proposed multi-marginal flow matching
methods and relate them to our novel ALIs. We begin by stressing a
distinction between our approach and all existing approaches: ALIs
are interpolants that follow distributions which match the
intermediate-time marginals, $q_{t_i}$, in distribution (see
\Cref{fig:ali}), while all existing multi-marginal interpolation
methods match the intermediate marginals pointwise.

The work of \cite{kapusniak2024metric, neklyudov2023computational} is closest
to our method, as they also learn nonlinear interpolants parameterised by a
time-dependent neural network. However, there are differences in both the
algorithms and the underlying motivations. Notably, neither of these approaches
explicitly approximates a divergence between the intermediate time marginals
and the interpolants, nor do they use adversarial training.
\cite{kapusniak2024metric} motivate their approach, metric flow matching (MFM),
by addressing distributions that are supported on a manifold with a metric
learning approach \citep{cox2008multidimensional,xing2002distance}. The metric
used in MFM can be chosen as either time-independent or time-dependent. The
former setting trivially contrasts to ALIs as our generator and discriminator
networks are time-dependent, while the latter case results in piecewise
interpolants (the interpolants are conditioned to pass through the marginal
samples), which again differs from our distribution-fitting approach.
Furthermore, our method does not rely on specifying a particular underlying
metric, and, importantly, we show experimentally in \Cref{sec:knot_exp} and
\Cref{sec:cell_tracking} how time-independent metrics prevent MFM from learning
valuable geodesics when the geometry of the underlying dynamics changes with
time.

\cite{rohbeck2025modeling, lee2025multi} propose an explicit
parametrisation of interpolants using splines. However,
these methods -- along with the ones discussed before
\citep{tong2023improving, neklyudov2023computational, kapusniak2024metric}
-- assume that the obtained interpolants should pass through the
available samples of marginal distributions.
In addition to that, spline-based interpolants
\citep{rohbeck2025modeling, lee2025multi} may scale poorly in high dimensions.

\section{Experiments}
\label{sec:experiments}
We conduct four different experiments. To showcase the flexibility of
our ALIs to solve multi-marginal problems
for large $K$, with noisy data, and where the data geometry changes
with time, we consider a toy experiment in
\Cref{sec:knot_exp} and an experiment with real cell tracking data in
\Cref{sec:cell_tracking}. In \Cref{sec:scrna_traj}
and \Cref{sec:st_exps}, we experiment with scRNA-seq and ST data,
respectively, where $K\in[4, 5]$. In all experiments,
we compare our method with the baselines in terms of earth mover's
distances (EMDs) and/or visual verifications.
Finally, note that all methods here are trained with the OT coupling
(see \Cref{sec:method}),  but that a comparison of
I-ALI-CFM and OT-ALI-CFM is provided in \Cref{app:cell_tracking}.
\begin{figure}[!t]
  \centering
\begin{minipage}{0.49\textwidth}
    \begin{subfigure}[t]{0.49\linewidth}
    \includegraphics[width=1\linewidth]{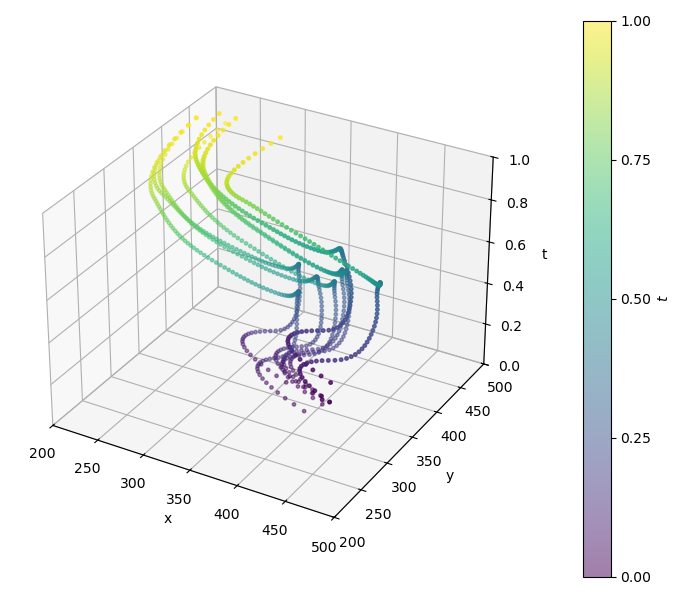}
    \caption{ALIs}
  \end{subfigure}\hfill
  \begin{subfigure}[t]{0.49\linewidth}
    \includegraphics[width=1\linewidth]{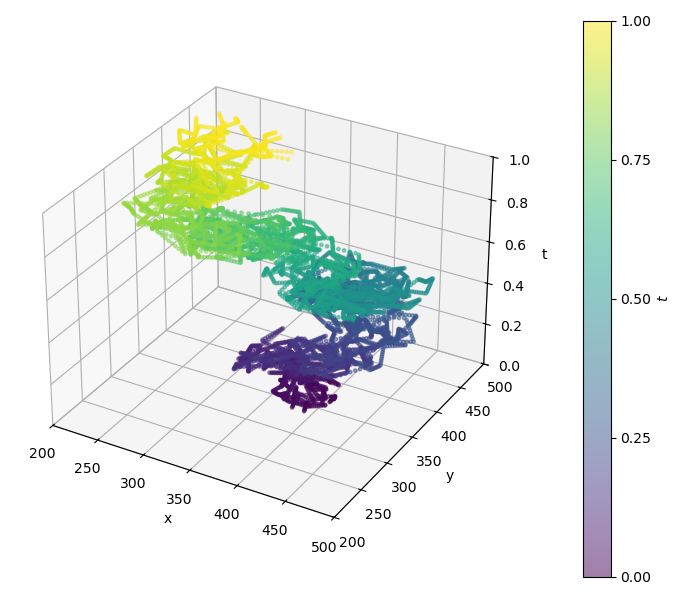}
    \caption{Piecewise linear interp.}
  \end{subfigure}


  \begin{subfigure}[t]{0.49\linewidth}
    \includegraphics[width=1\linewidth]{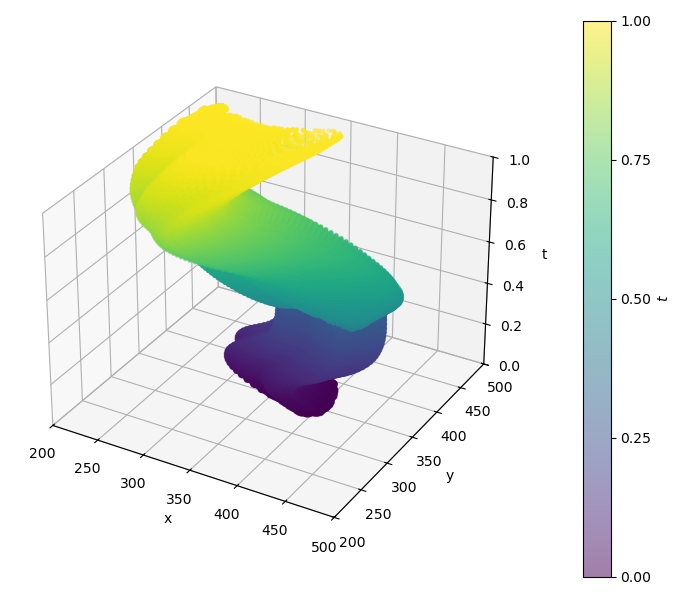}
    \caption{OT-ALI-CFM}
  \end{subfigure}\hfill
  \begin{subfigure}[t]{0.49\linewidth}
    \includegraphics[width=1\linewidth]{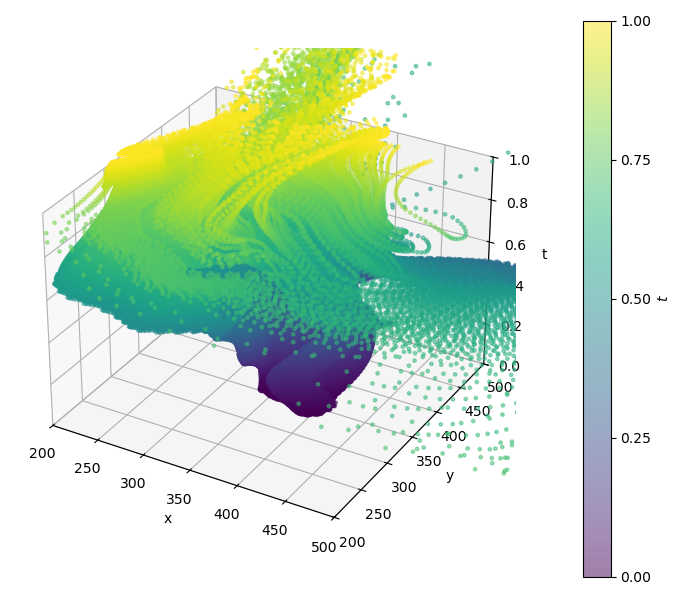}
    \caption{OT-CFM}
  \end{subfigure}
\end{minipage}
    \hfill
\begin{minipage}{0.49\textwidth}
        \begin{subfigure}[t]{0.49\linewidth}
    \includegraphics[width=0.83\linewidth]{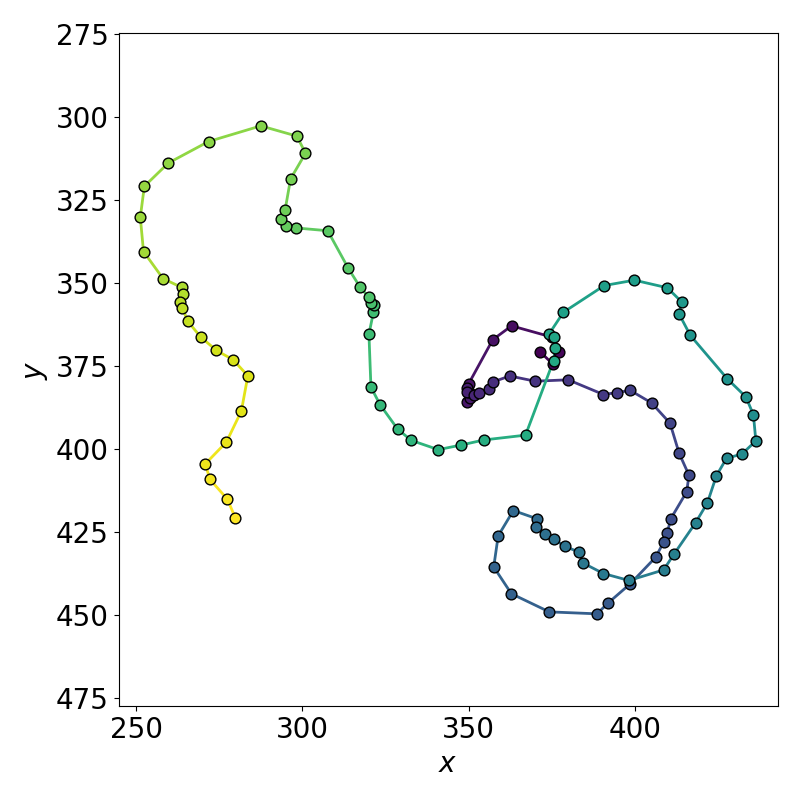}
    \caption{All data}
    \label{fig:data_centroid}
  \end{subfigure}\hfill
  \begin{subfigure}[t]{0.49\linewidth}
    \includegraphics[width=0.9\linewidth]{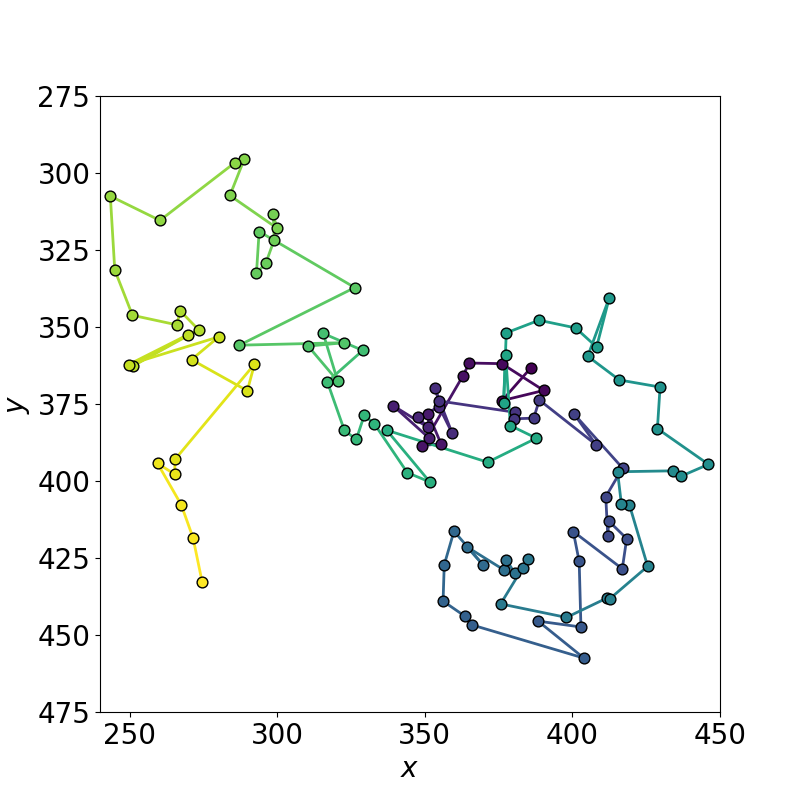}
    \caption{Subset of data}
  \end{subfigure}


  \begin{subfigure}[t]{0.49\linewidth}
    \includegraphics[width=0.9\linewidth]{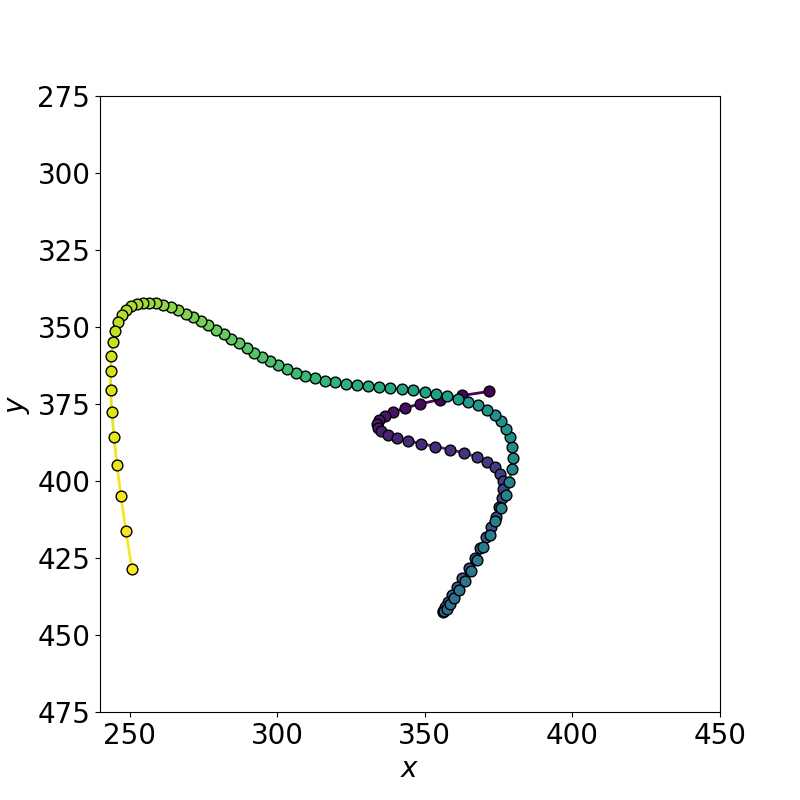}
    \caption{OT-ALI-CFM}
  \end{subfigure}\hfill
  \begin{subfigure}[t]{0.49\linewidth}
    \includegraphics[width=0.9\linewidth]{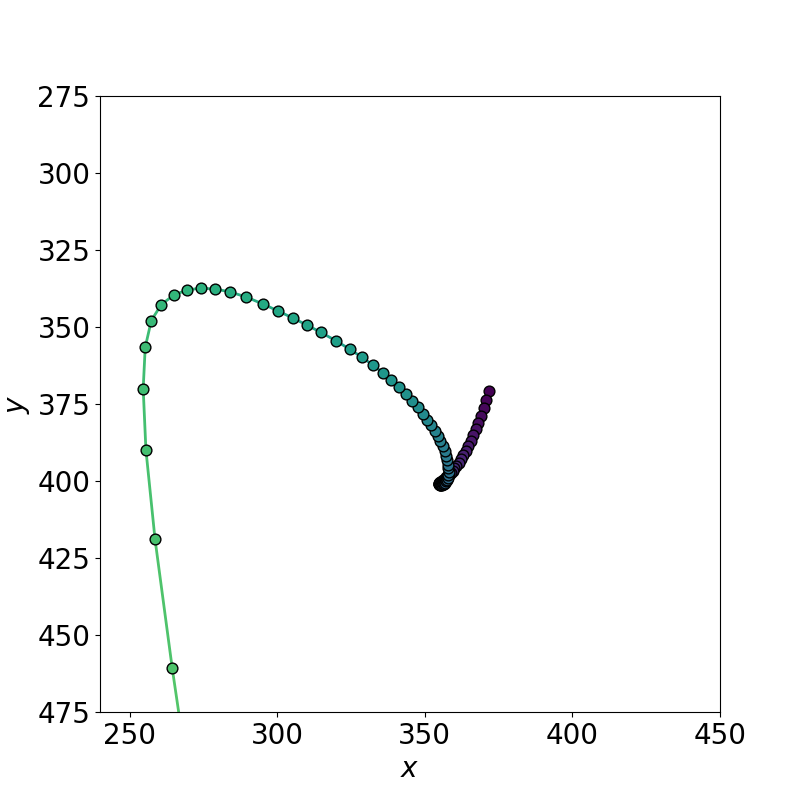}
    \caption{OT-CFM}
    \label{fig:cell_centroids_ot_cfm}
  \end{subfigure}
\end{minipage}
    \vspace{-0.5\baselineskip}
  \caption{\textbf{(a)} and \textbf{(b)} show ten ALIs and piecewise linear interpolants, respectively, based on the subsampled cell tracking data, while \textbf{(c)} and \textbf{(d)} depict the resulting OT-ALI-CFM and OT-CFM vector fields. In the 2D figures we plot the centroids (one per frame) of \textbf{(e)} all of the segmentation data, \textbf{(f)} the subset training data, \textbf{(g)} the OT-ALI-CFM trajectories and \textbf{(h)} the OT-CFM trajectories. Note that the OT-CFM trajectories diverge in \textbf{(d)} and \textbf{(h)}.}
  \label{fig:cell_tracking_trajectories}
\end{figure}

\subsection{Synthetic data}
\label{sec:knot_exp}
First, we showcase the flexibility of our ALI-CFM method on synthetic
data: a sequence of 1,200 marginal
distributions centred along a knot (\Cref{fig:figure1}). We obtain 10
samples from each marginal distribution.
The experiment shows that ALI-CFM is the only method capable of
accurately capturing the time-dependent
geometry of the `knot' distribution (formalised in \Cref{app:knot}).
For OT-MFM, we considered the time-dependent
LAND metric, where a sequence of LAND metrics is constructed from the
data at $t_i$ and $t_{i+1}$, and the
interpolants are learnt using all the data. More experimental details
are provided in \Cref{app:knot_exp}.

Notably, the piecewise linear interpolant \citep{tong2023improving}, the cubic
spline interpolants \citep{rohbeck2025modeling} and OT-MFM with time-dependent
LAND metrics are non-smooth, which results in
the corresponding CFM objective having high gradient variance. This was not the case when training OT-ALI-CFM on the
smooth ALI interpolants. In \Cref{fig:interpolant-comparison} we visually
compare the smoothness of ALIs and OT-MFM interpolants for different choices of
$K$, demonstrating that MFM scales poorly with $K$, while our ALIs consistently
produce smooth and accurate interpolants.

\subsection{Learning spatio-temporal cellular dynamics with cell tracking data}
\label{sec:cell_tracking}
Encouraged by our results in \Cref{fig:figure1}, we aim to learn the
spatio-temporal dynamics
of the positions of a cell. That is, given segmentation masks of a
glioblastoma-astrocytoma
U373 cell collected in 115 times steps/consecutive frames, provided
by The Cell Tracking Challenge
\citep{mavska2023cell}, we consider learning vector fields that, when
integrated, produce
trajectories that follow the tracked cell movements. This is
challenging as the shape of the cell
contracts and expands while it moves along a loopy path
(\Cref{fig:data_centroid}).

To demonstrate ALIs' capacity to handle noisy datasets, we independently
subsample ten segmentation coordinates per time stamp. Based on the subset, we
learn both interpolants and CFMs. Then, we use all $x_0$ samples when
integrating the vector field, i.e., we push all $x_0$ samples in the dataset
through the learnt vector fields with 115 integration steps (same as the number
of frames). Given the similar performances of OT-CFM and OT-MMFM in
\Cref{fig:figure1}, we compare OT-ALI-CFM to OT-CFM and OT-MFM in this
experiment. Note that, here, we implemented two versions of OT-MFM using the
time-independent and time-dependent LAND metric. The latter setting is the same
as in \Cref{sec:knot_exp}, while the former mimics the LiDAR experiment setup
in \cite{kapusniak2024metric}, where the metric is inferred from all the data,
while the interpolants are learnt using only samples from $q_0$ and $q_1$.

\begin{figure}[!t]
  \centering
  \begin{minipage}{0.15\textwidth}
    Training data
    \end{minipage}
    \hfill
    \begin{minipage}{0.84\textwidth}
        \begin{subfigure}[t]{0.9\linewidth}
    \includegraphics[width=\linewidth]{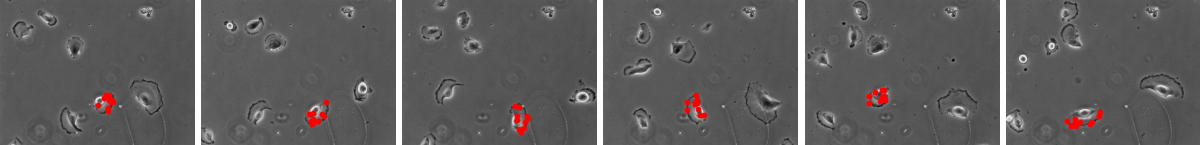}
  \end{subfigure}
    \end{minipage}


\begin{minipage}{0.15\textwidth}
    OT-ALI-CFM
    \end{minipage}
    \hfill
    \begin{minipage}{0.84\textwidth}
        \begin{subfigure}[t]{0.9\linewidth}
    \includegraphics[width=\linewidth]{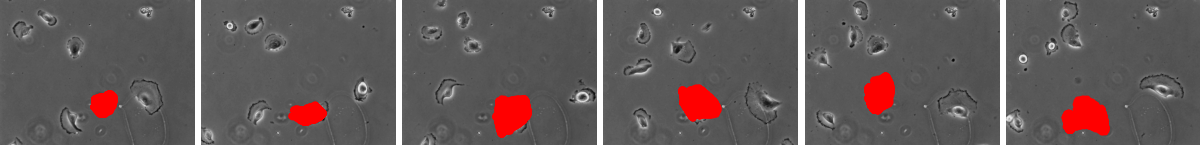}
  \end{subfigure}
    \end{minipage}


\begin{minipage}{0.15\textwidth}
    OT-CFM
    \end{minipage}
    \hfill
    \begin{minipage}{0.84\textwidth}
        \begin{subfigure}[t]{0.9\linewidth}
    \includegraphics[width=\linewidth]{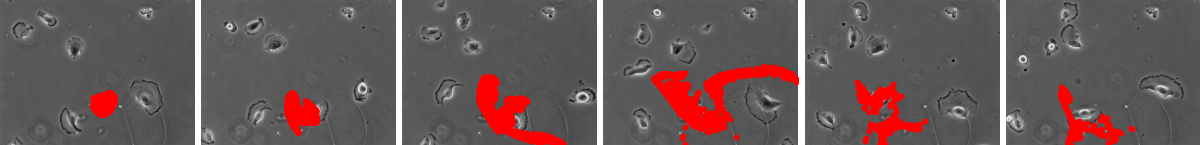}
  \end{subfigure}
    \end{minipage}
\begin{minipage}{0.15\textwidth}
    OT-MFM
    \end{minipage}
    \hfill
    \begin{minipage}{0.84\textwidth}
        \begin{subfigure}[t]{0.9\linewidth}
    \includegraphics[width=\linewidth]{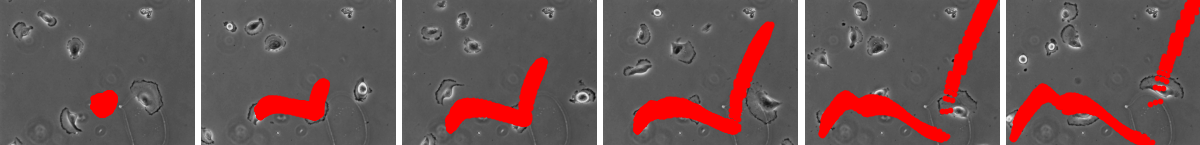}
  \end{subfigure}
    \end{minipage}
    \vspace{-0.5\baselineskip}
  \caption{From left to right we visualize frame 1, 25, 50, 75, 100 and 115 in the sequence of microscopy images. From top to bottom we overlay the images with the subsampled training data (10 samples per frame), and all segmentation samples $x_0$ pushed to the frame-specific times via the marginalised vector fields of OT-ALI-CFM, OT-CFM and OT-MFM. In concordance with the centroid plot in \Cref{fig:cell_centroids_ot_cfm}, the trajectories from OT-CFM eventually diverge and become inaccurate. Meanwhile the time-independent metric in OT-MFM results in trajectories that do not capture the cell positions' temporal dependence.}
  \label{fig:cell_tracking_masks}
\end{figure}

Inspecting the resulting interpolants in \Cref{fig:cell_tracking_trajectories},
ALIs provide a smooth map between $(x_0, x_1)$ pairs while the piecewise linear
interpolants, again, introduce instability reflected as non-smoothness in the marginalised trajectories. %
Therefore, we could not train an OT-CFM without the integrated vector fields
diverging, whereas fitting an OT-ALI-CFM was straightforward.
In \Cref{fig:cell_tracking_masks}, we overlay the training data and the
inferred CFM trajectories on the microscopy images, visualising also the
inability of time-independent OT-MFM to learn accurate dynamics when the data
geometry changes with time. The plots in \Cref{fig:mfm_cell_tracking} support
the latter observation. Meanwhile, the time-dependent OT-MFM produces similar
interpolants to OT-CFM, which are not smooth due to their
piecewise nature, causing the vector field integration to diverge
(\Cref{fig:time_dependent_otmfm_3d}). We constrained the analysis here to
visual evaluations as the results in the mentioned figures are conclusive.

To produce \Cref{fig:cell_tracking_trajectories}, we trained the CFM
nets for 50,000 iterations with a learning rate of $10^{-3}$
and minibatches of size 128. The CFM nets were 3-hidden-layer MLPs
with 256 units in each layer and SELU activations. ALI was
regularised with $\lambda=1$ and  \eqref{eq:reg}. More details on
data and hyperparameters are provided in \ref{app:cell_tracking}.

\subsection{Single-cell trajectory inference}
\label{sec:scrna_traj}
As another real-world example, we experiment with time-series
single-cell data, predicting the gene expression levels of cell
populations that evolve over time. The samples from the $K$
consecutive populations are naturally unpaired, and the dynamics of
the gene expressions are unknown. We follow preceding works
\citep{tong2023improving} on predicting the marginal distributions of
held-out intermediate-time data, measuring accuracy via the
earth-mover distance (EMD) between the CFM trajectories and the
held-out data at the corresponding time stamp. That is, using the
remaining data, we train ALIs by optimising the objective in
\eqref{eq:ali_objective} and then learn ALI-CFMs by minimising
\eqref{eq:cfm}. In \Cref{tab:5D_experiments,tab:highdim}, we compare
the performance of OT-ALI-CFM to other CFM methods on Embryoid body
(EB; \cite{moon2019visualizing}) data, and on Cite-seq (Cite) and
Multiome (Multi) data from \cite{lance2022multimodal}. The
preprocessing pipeline of single-cell data closely follows that of
\citet{kapusniak2024metric}. The only difference is that we normalise
the data to ensure stability during training and denormalise only for
metric computations. More details are given in
\Cref{app:sc_experiments}, where we also share an ablation study of
the effect of our regularisers on the trajectory curvature and on the EMDs.

Although our algorithm is on par with existing baselines, we believe that the
nature of adversarial training makes it difficult to completely outperform
OT-MFM. Since our adversarially learnt interpolant matches the points at each
time step in a distribution-matching sense, it might lose in pointwise metrics
to methods that are trained to overfit to the given data.

\begin{table}[!t]
  \centering
  \begin{minipage}{0.3\textwidth}
    \caption{
      Trajectory inference on 5D PCA scRNA-seq data. Accuracy
      measured in EMD (smaller is better) w.r.t.
      the left-out marginal distributions, averaged over five
    independent runs.    \label{tab:5D_experiments}}
  \end{minipage}
  \hfill
  \begin{minipage}{0.69\textwidth}
    \resizebox{\linewidth}{!}{
      \begin{tabular}{@{}lccc}
        \toprule
        Algorithm $\downarrow$
        Dataset $\rightarrow$   & Cite             & EB
        & Multi            \\
        \midrule
        I-CFM                   & 1.236\std{0.050} & 1.156\std{0.42}
        & 1.150\std{0.091} \\
        OT-CFM                  & 1.142\std{0.085} & 0.809\std{0.16}
        & 0.975\std{0.045} \\
        OT-MFM                  & 0.793\std{0.019} & 0.711\std{0.050} &
        0.890\std{0.123}
        \\
        I-MMFM (Cubic splines)  & 2.068\std{0.390} & 4.740\std{0.650}
        & 1.528\std{0.040} \\
        OT-MMFM (Cubic splines) & 1.099\std{0.043} & 3.530\std{0.194}
        & 1.807\std{0.085} \\
        \midrule
        OT-ALI-CFM (ours)       & 0.910\std{0.024} & 0.742\std{0.022}
        &
        0.925\std{0.018}%
        \\
        \bottomrule
      \end{tabular}
    }
  \end{minipage}
\end{table}

\begin{table}[t]
  \centering
  \begin{minipage}{0.3\linewidth}
    \caption{
      Trajectory inference on 50D and 100D PCA scRNA-seq data.
      The accuracies here are measured in the same way as in
      \Cref{tab:5D_experiments}.\label{tab:highdim}
    }
  \end{minipage}
  \hfill
  \begin{minipage}{0.69\linewidth}
    \resizebox{\linewidth}{!}{
      \begin{tabular}{@{}llllllll}
        \toprule
        Dim. $\rightarrow$                              &
        \multicolumn{2}{c}{50} & \multicolumn{2}{c}{100}
        \\
        \cmidrule(lr){2-3}\cmidrule(lr){4-5}
        Algorithm\ $\downarrow$ | Dataset $\rightarrow$ &
        \multicolumn1c{Cite}   & \multicolumn1c{Multi}   &
        \multicolumn1c{Cite} & \multicolumn1c{Multi} \\
        \midrule
        I-CFM                                           &
        42.478\std{0.930}      & 51.098\std{0.340}       &
        49.929\std{0.391}    & 57.801\std{0.365}     \\
        OT-CFM                                          &
        38.367\std{0.295}      & 47.205\std{0.184}       &
        45.148\std{0.207}    & 54.630\std{0.456}     \\
        I-MFM                                           &
        41.172\std{0.269}      & 48.415\std{0.793}       &
        46.339\std{0.618}    &
        53.667\std{0.768}
        \\ OT-MFM            & 36.471\std{0.480}
        & 45.879\std{0.438} &
        42.232\std{0.249}                               &
        51.169\std{0.523}
        \\ \midrule OT-ALI-CFM (ours) &
        41.449\std{0.942}                               &
        46.454\std{0.538}      & 48.496\std{0.814}       & 54.554\std{0.832}
        \\ \bottomrule
      \end{tabular}
    }
  \end{minipage}
\end{table}

\newpage

\subsection{Tumour coordinate inference with spatial transcriptomics data}
\label{sec:st_exps}
\begin{wraptable}{r}{0.35\linewidth}
  \centering
  \caption{Average EMD on ST data using learnt vector fields.}
  \label{tab:ST_tab}
  \vspace*{-1em}
  \begin{tabular}{@{}lc}
    \toprule
    Method     & EMD $(\downarrow)$       \\
    \midrule
    OT-CFM     & 109.76\std{9.98}         \\
    OT-MMFM    & 109.17\std{9.82}         \\
    OT-MFM     & 183.88\std{53.92}        \\
    \midrule
    OT-ALI-CFM & \textbf{98.91\std{2.03}} \\
    \bottomrule
  \end{tabular}

\end{wraptable}
Understanding cancer evolution is a key problem to solve for accurate
tumor analysis and effective
treatment development \citep{greaves2012clonal}. ST
\citep{staahl2016visualization}
has recently enabled studies of spatial tumor dynamics
\citep{shafighi2024integrative}.
In an ST platform, a thin section of a tissue is placed on a slide
with coarsely distributed features,
where cells in the tissue are sequenced, while maintaining the
spatial locations of the cells.
As such, given a single tissue section, any ST platform returns a 2D
distribution of gene expression data,
often accompanied by an hematoxylin and eosin (H\&E) stained image of
the section.
Multiple sections are typically processed, allowing for informative
analyses of the spatial organisation of cancer cells.

We consider the breast cancer dataset published in \cite{mo2024tumour}, using
$K=4$ sections, each processed with the Visium 10x Genomics platform. There are
more than 1,000 samples in each section, each $x_t\in\mathbb{R}^2$, and the
number of samples vary with $t$. We seek to model the distribution of tumour
regions in an unobserved tissue section. As far as we are aware, this task has
not previously been attempted, however it is well-suited to being tackled using
FM as the samples (coordinates) in each section are naturally unpaired.%

\begin{figure}[!t]
  \centering
  \vspace*{-1em}
  \includegraphics[width=1\linewidth]{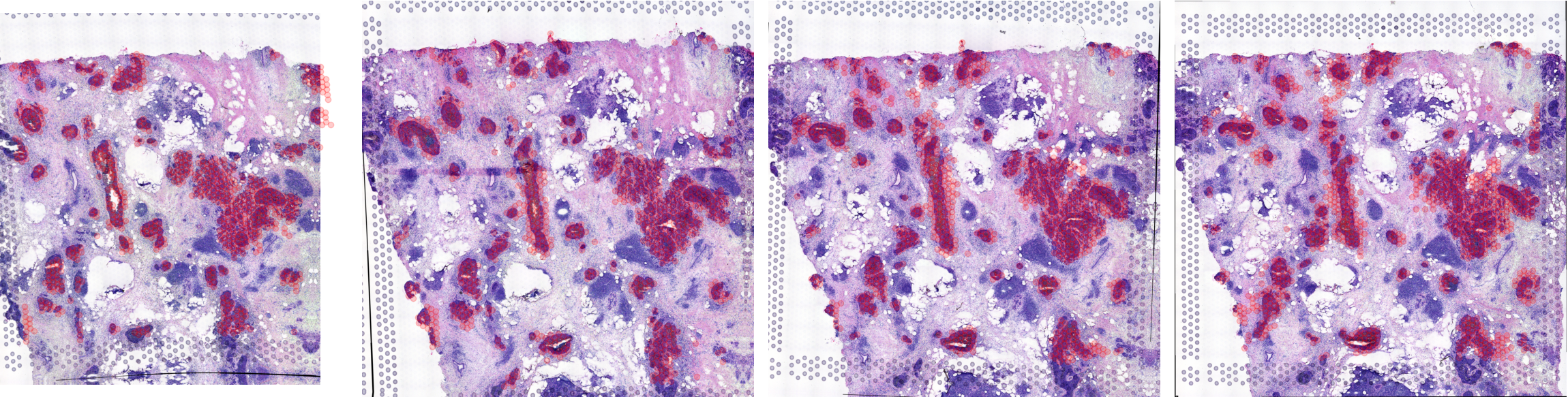}
  \caption{The aligned breast cancer H\&E-stained images with
    overlaid scatter plots of tumor annotated
  coordinates (in light red).}
  \label{fig:BRCA_data}
  \vspace*{-1em}
\end{figure}

The four sections are spatially ordered along the $z$-axis, and the authors of
the dataset have assigned the coordinates in each section with a tumour purity
score based on the measured gene expressions. We classified all coordinates
with a score above 0.8 as belonging to a tumour region\footnote{The threshold
  was chosen based on visual agreement between the tumour-classified coordinates
  and the tumour regions in the corresponding H\&E-stained images
  (darker regions
typically imply higher density of tumour cells).} and aligned the sections
(details in \Cref{app:ST_preprocessing}), which enabled us to consider a common
coordinate system across sections. We visualise the preprocessed data in
\Cref{fig:BRCA_data} and the raw images in \Cref{fig:BRCA_raw_imgs}.

\begin{figure}[ht]
  \centering
  \begin{subfigure}[t]{0.19\textwidth}
    \centering
    \includegraphics[width=\textwidth]{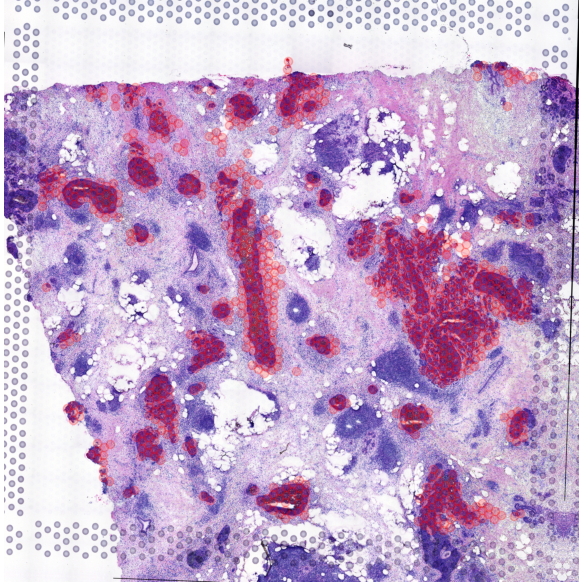}
    \caption{}
  \end{subfigure}\hfill
  \begin{subfigure}[t]{0.19\textwidth}
    \centering
    \includegraphics[width=\textwidth]{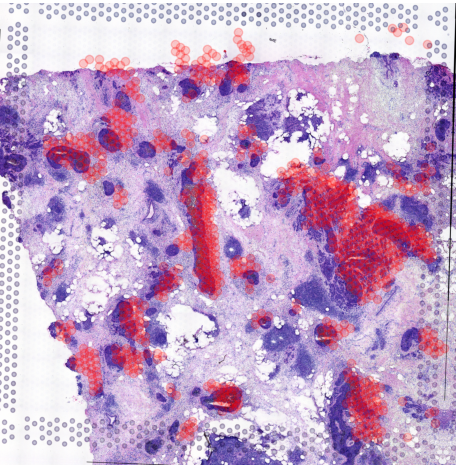}
    \caption{}
  \end{subfigure}\hfill
  \begin{subfigure}[t]{0.19\textwidth}
    \centering
    \includegraphics[width=\textwidth]{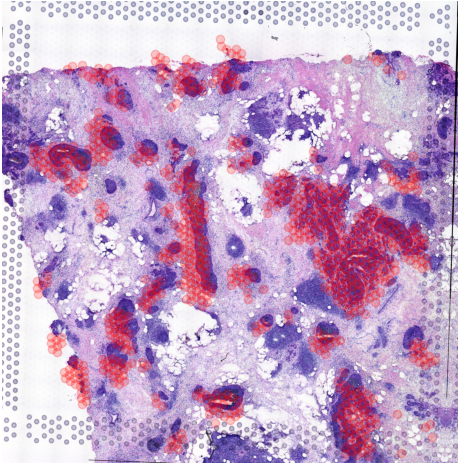}
    \caption{}
  \end{subfigure}\hfill
  \begin{subfigure}[t]{0.19\textwidth}
    \centering
    \includegraphics[width=\textwidth]{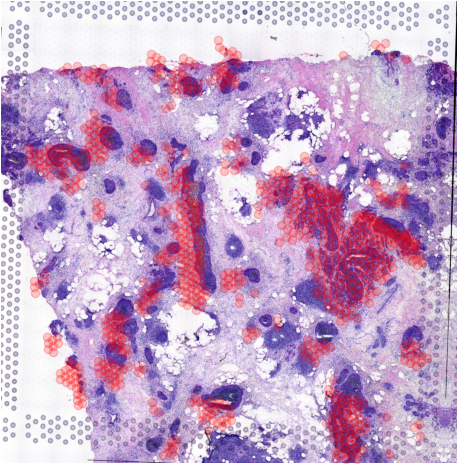}
    \caption{}
  \end{subfigure}
  \begin{subfigure}[t]{0.19\textwidth}
    \centering
    \includegraphics[width=\textwidth]{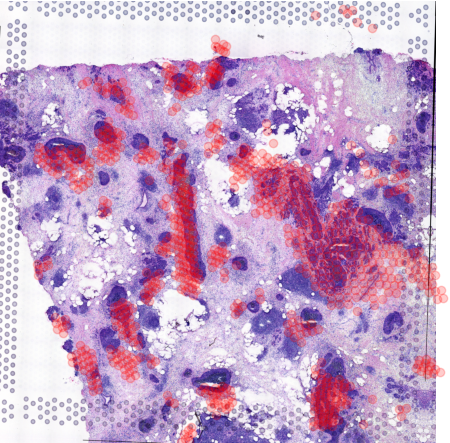}
    \caption{}
  \end{subfigure}
  \caption{
    H\&E-stained histology image of the third breast cancer tissue
    section overlayed with \textbf{(a)} true
    tumour annotated coordinates, and CFM trajectories integrated to
    $t=2/3$ using \textbf{(b)} OT-ALI-CFM,
    \textbf{(c)} OT-CFM, \textbf{(d)} OT-MMFM and \textbf{(e)}
    OT-MFM. OT-ALI-CFM achieves the lowest EMD
    score w.r.t. to held-out coordinates in \textbf{(a)} (\Cref{tab:ST_tab}).
  }
  \label{fig:inferred_BRCA}
\end{figure}

Following the notational convention, we let pseudo-times $t\in[0,1]$ correspond
to normalised $z$-coordinates, such that $q_0$ is the distribution of
coordinate samples in the bottom section, and so on. We assume equal spacing
between the sections, which is a reasonable assumption based on the
documentation in \cite{mo2024tumour}. We experiment with leave-one-out
interpolation, omitting either section two ($t=1/3$) or three ($t=2/3$) from
the training data. We then compare the EMD between the held-out data and the
integrated vector fields. The EMD results are shown in \Cref{tab:ST_tab}, and a
visualisation is provided in \Cref{fig:inferred_BRCA}.

This dataset is challenging as the marginal distributions are highly
multimodal, with modes that are not present in all sections. Also, tissues can
shrink when stored before processing which makes alignment challenging
(\Cref{app:ST_preprocessing}; \cite{zeira2022alignment}). These challenges
explain the OT-CFM and OT-MMFM results, while we believe that this makes the
time-dependent LAND metric in OT-MFM inaccurate.




Interestingly, although GAN optimisation is considered sensitive to multimodal
data distributions \citep{arjovsky2017wasserstein, huanggan2024ganisdead}, we
found that our regulariser in \eqref{eq:piecewise_reg} along with the
Markov-chained OT coupling made training stable, with clearly impressive
results.

All CFMs were trained using MLPs with three hidden layers, 128 units per layer
and SELU activation functions. The CFMs were trained for 10,000 epochs with a
batch size of 128. $D_\gamma$ and $G_\phi$ were small MLP nets with two hidden
layers, 64 units per layer and ELU activations. In the ALI objective
\eqref{eq:ali_objective} we used the piecewise-linear interpolant regulariser
in \eqref{eq:piecewise_reg}. See \Cref{app:st} for more details.

\section{Discussion}
We propose ALI-CFM, a novel approach to learning interpolants in multi-marginal
flow matching. ALI is a conceptually novel way to learn interpolations by
matching the distributions of intermediate marginals, as opposed to the strict
pointwise assignment used in all existing methods. We have demonstrated the
superior capabilities of our method to fit complex dynamics with noisy samples
from hundreds (\Cref{sec:cell_tracking}) or thousands (\Cref{sec:knot_exp}) of
marginal distributions -- no other FM algorithm managed to demonstrate a
comparable performance on this task. Moreover, across datasets and dimensions,
our ALI-CFM performs on par with the state-of-the-art methods on single-cell
trajectory inference.

Our interpolants are learnt using an adversarial learning objective, which fits
well to the density-free problem setting in flow matching. Although GANs are
challenging to train, we have proposed regularising terms to the objective
that, apart from guaranteeing unique interpolants, empirically stabilise
training.

In fact, compared to baselines (see the final paragraph in
\Cref{app:knot_exp}), our adversarial training can reduce the complexity of
training accurate interpolants in flow matching. Additionally, there is a rich
literature on GAN optimisation, e.g., regarding alternative objectives
\citep{arjovsky2017wasserstein, jolicoeur2018relativistic, sun2020towards,
huanggan2024ganisdead}.

\newpage
\section*{Acknowledgements}

The authors are grateful to Alex Tong for helpful discussions about single-cell
data preprocessing and methods from prior work. The authors also thank Hosein
Toosi for providing valuable guidance regarding the preprocessing of the ST
data.

The work of VE is supported by the Advanced Research and Invention Agency
(ARIA). JL acknowledges support from the Swedish Research Council (project ID
2022-03516\_VR). ESW acknowledges support from the CIFAR Learning in Machines
and Brains programme.

This work was enabled by the computational resources of the Edinburgh
International Data Facility (EIDF) with funding from the Edinburgh Generative
AI Laboratory (GAIL).

\newpage
\appendix

\section{Proof of uniqueness theorems}
\label{app:proof}
Here we restate the theorems from \Cref{sec:regularisers} and provide
their proofs.
\printProofs

\section{Knot distribution}\label{app:knot}

We let $t\in [0,t_\text{max}]$, ensure that the number of marginals,
$K$, is a multiple of three, and set
\begin{equation}
  \tilde{t} = 3 \frac{t}{t_\text{max}} - 1.5.
\end{equation}
We then partition $\tilde{t}$ into three equally sized intervals,
\[
  \mathcal{I}_1 = \tilde{t}_{1:K/3}, \quad \mathcal{I}_2 =
  \tilde{t}_{K/3+1:2K/3} \quad \mathcal{I}_3 = \tilde{t}_{2K/3+1:K},
\]
and sample $X$ and $Y$ coordinates as follows:
\begin{align}
  \begin{bmatrix} X(t) \\ Y(t)
  \end{bmatrix} \sim
  \mathcal{N}\left(
    \begin{bmatrix} \mu_X(t) \\ \mu_Y(t)
    \end{bmatrix}, \sigma^2 I_2
  \right)
\end{align}
with $\sigma = 0.1$ and
\begin{subequations}
  \begin{align}
    \mu_X(t) &=
    \begin{cases}
      3(t + 0.5)            & \text{if } t \in \mathcal{I}_1 \\
      \cos(2\pi(t - 0.75))  & \text{if } t \in \mathcal{I}_2 \\
      3(t - 0.5)            & \text{if } t \in \mathcal{I}_3
    \end{cases} \\[1.5ex]
    \mu_Y(t) &=
    \begin{cases}
      -0.5 \tanh(5(t + 1)) + 0.5 & \text{if }  t \in \mathcal{I}_1 \\
      \sin(2\pi(t - 0.75))       & \text{if }  t \in \mathcal{I}_2 \\
      0.5 \tanh(5(t - 1)) + 0.5  & \text{if }  t \in \mathcal{I}_3
    \end{cases}
  \end{align}
\end{subequations}
Finally, we divide $t$ by three and collect ten samples from the
bivariate distribution of ($X(t), Y(t)$) at
each of the $K$ time stamps.

\section{Preprocessing the spatial transcriptomics data}
\label{app:ST_preprocessing}
\begin{figure}
  \centering
  \begin{subfigure}[t]{0.24\linewidth}
    \includegraphics[width=1\linewidth]{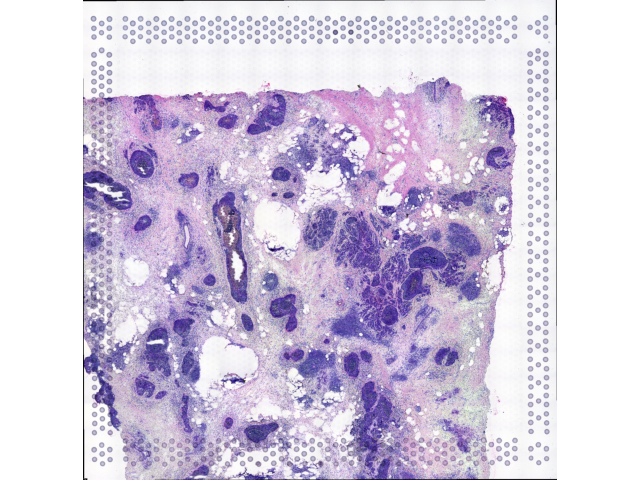}
    \caption{}
  \end{subfigure}
  \begin{subfigure}[t]{0.24\linewidth}
    \includegraphics[width=1\linewidth]{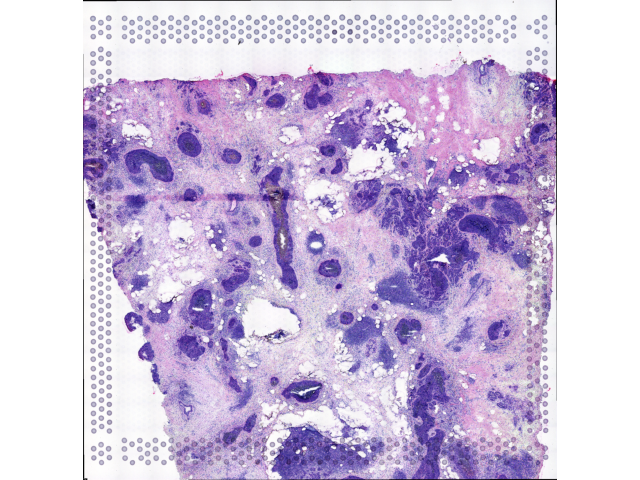}
    \caption{}
  \end{subfigure}
  \begin{subfigure}[t]{0.24\linewidth}
    \includegraphics[width=1\linewidth]{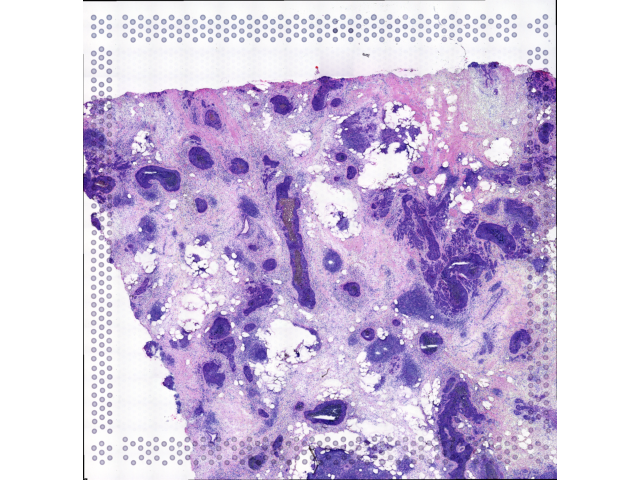}
    \caption{}
  \end{subfigure}
  \begin{subfigure}[t]{0.24\linewidth}
    \includegraphics[width=1\linewidth]{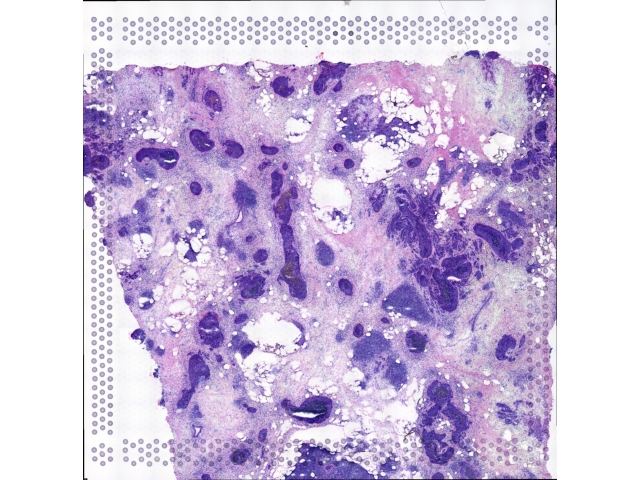}
    \caption{}
    \label{fig:brca_u5}
  \end{subfigure}
  \caption{The raw H\&E-stained images of the four tissue sections
  provided by \cite{mo2024tumour}.}
  \label{fig:BRCA_raw_imgs}
\end{figure}

A sequence of spatial transcriptomics tissue sections (e.g., in
\Cref{fig:BRCA_raw_imgs}) is not naturally aligned
since each slice is cut separately, for instance, at slightly
different orientations. Small physical distortions
occur during sectioning, staining, storing (the tissues are typically
frozen post-sectioning), and imaging, causing
stretching, folding, or rotation relative to neighbouring sections.
In fact, practitioners manually place the tissues
on a slide.
As a result, the spatial coordinates reported for each section exist
in their own local coordinate system and cannot
be directly mapped to the spots of other sections without an
alignment preprocessing step. We are particularly interested
in aligning the spatial coordinates of the spots (the spots, or
features, contain the RNA expression).

We aligned the spots by first visually aligning the raw H\&E-stained images
shown in \Cref{fig:BRCA_raw_imgs} using the BigWarp plugin
\citep{bogovic2016robust} in Fiji \citep{schindelin2012fiji}, an open-source
biological-image analysis software. After downscaling the H\&E-stained images
by a factor of 1/10, we pairwise aligned all images to the fourth section
(\Cref{fig:brca_u5}). To align a pair of images in BigWarp one manually chooses
landmarks which the software then utilises to nonlinearly warp the source image
to the target image (again, \Cref{fig:brca_u5} is the target image in all image
pairs). The above accurately aligned the pixel coordinates of the source image
to the target, which was clear from visual inspection.

Finally, we transformed the spot coordinates in the source section to the
coordinate system of the target section. In particular, the nonlinear
transformation used above (thin plate spline transformation) provided a mapping
from the source pixel coordinates to the target pixel coordinate system. We
found that feeding these maps to a
\texttt{scipy.interpolate.RegularGridInterpolator} object to transform the
source spot coordinates accurately aligns the spot coordinates with the warped
source images. The result of our alignment preprocessing is shown in
\Cref{fig:BRCA_data}.

\section{Additional experimental details and results}\label{app:exps}

\subsection{Knot experiment}
\label{app:knot_exp}
Our proposed method is trained using two-layer MLPs with $128$ hidden
units and ELU activation function
for both $f_\phi$ and $D_\gamma$. We use $\lambda=1$ in
\eqref{eq:ali_objective}, the linear reference
regulariser from \eqref{eq:reg}, and we train the interpolant for
$50,000$ epochs with a batch size of $128$.
The nets $f_\phi$ and $D_\gamma$ are optimised using separate Adam
optimisers with a learning rate of $10^{-3}$.
The flow matching networks are parameterised as two-layer MLPs with
$32$ hidden units and SELU activations in
every experiment, and the Adam optimiser \citep{kingma2014adam} is
used for training. OT-ALI-CFM, OT-CFM and
OT-MMFM were trained for $30000$ epochs with a learning rate of
$10^{-4}$. See below for OT-MFM details.

\paragraph{OT-MFM hyperparameters with time-independent LAND metric}
We train OT-MFM for $10$, $1,000$, and $40,000$ epochs with Adam optimiser and
learning rate $10^{-4}$ for both the interpolant and flow matching networks. We
experimented with other learning rates; however, we observed no difference or a
degraded performance. We kept the default hyperparameters from the MFM codebase
\citep{kapusniak2024metric} with the LAND metric. Following the net parameter
choices for ALI on this data, the interpolant network has two hidden layers
with $128$ units, and the vector field net has two layers and $64$ hidden
units. All networks have ELU activations. The batch size is set to $128$.

\begin{figure}
  \centering

  \begin{tabular}{@{} *{7}{c@{}}}
    & $K=30$ & $K=60$ & $K=120$ & $K=240$ & $K=480$ & $K=1200$ \\
    \raisebox{1.1cm}{\scriptsize \rotatebox[origin=c]{90}{LAND-MFM}} &
    \includegraphics[width=0.163\linewidth]{
      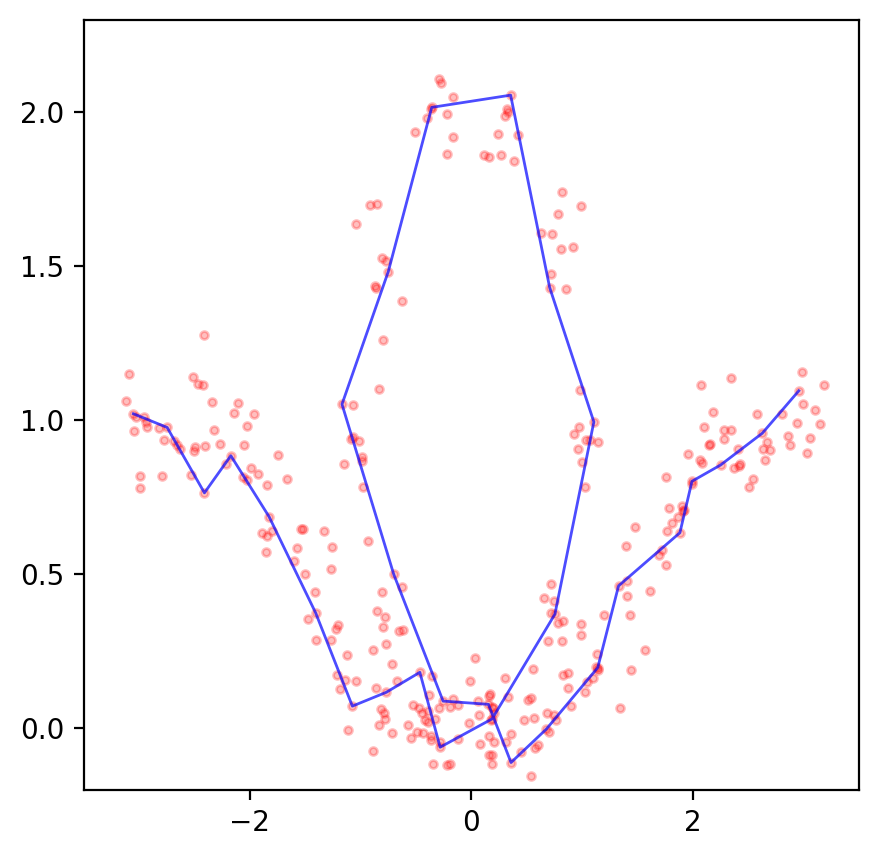
    }                                                                &
    \includegraphics[width=0.163\linewidth]{
      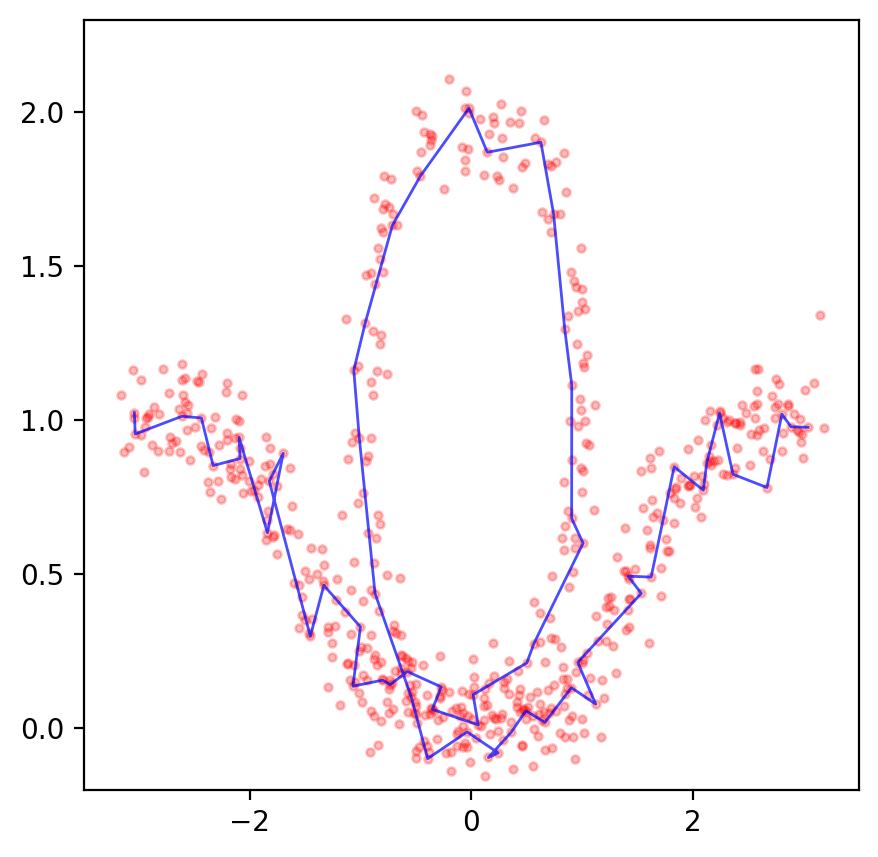
    }                                                                &
    \includegraphics[width=0.163\linewidth]{
      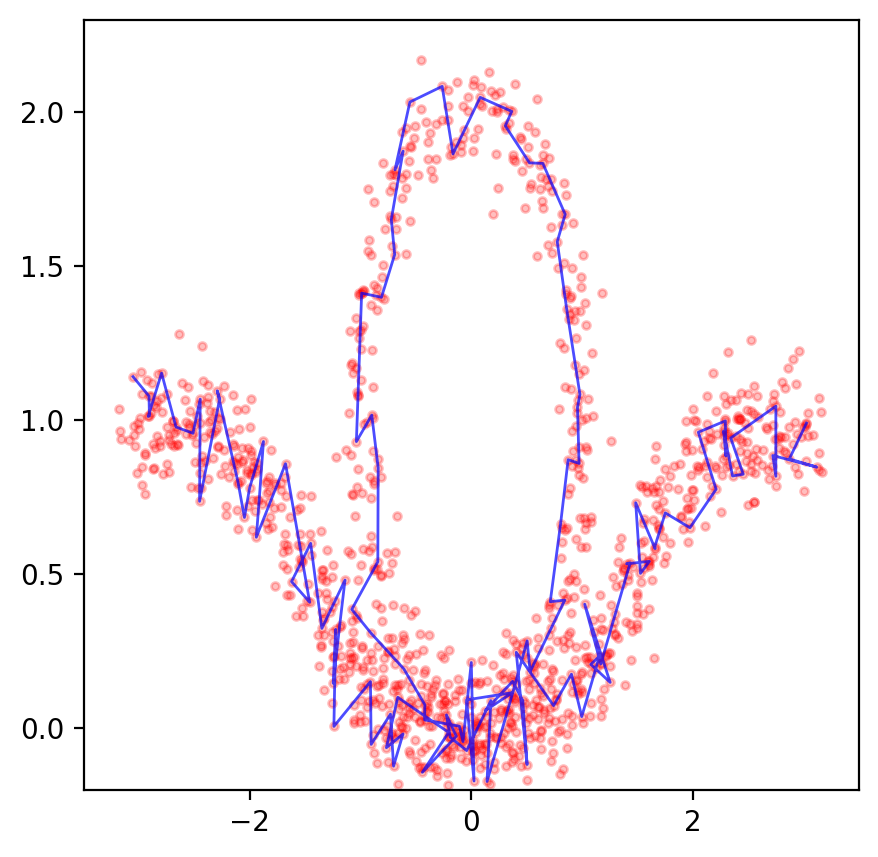
    }                                                                &
    \includegraphics[width=0.163\linewidth]{
      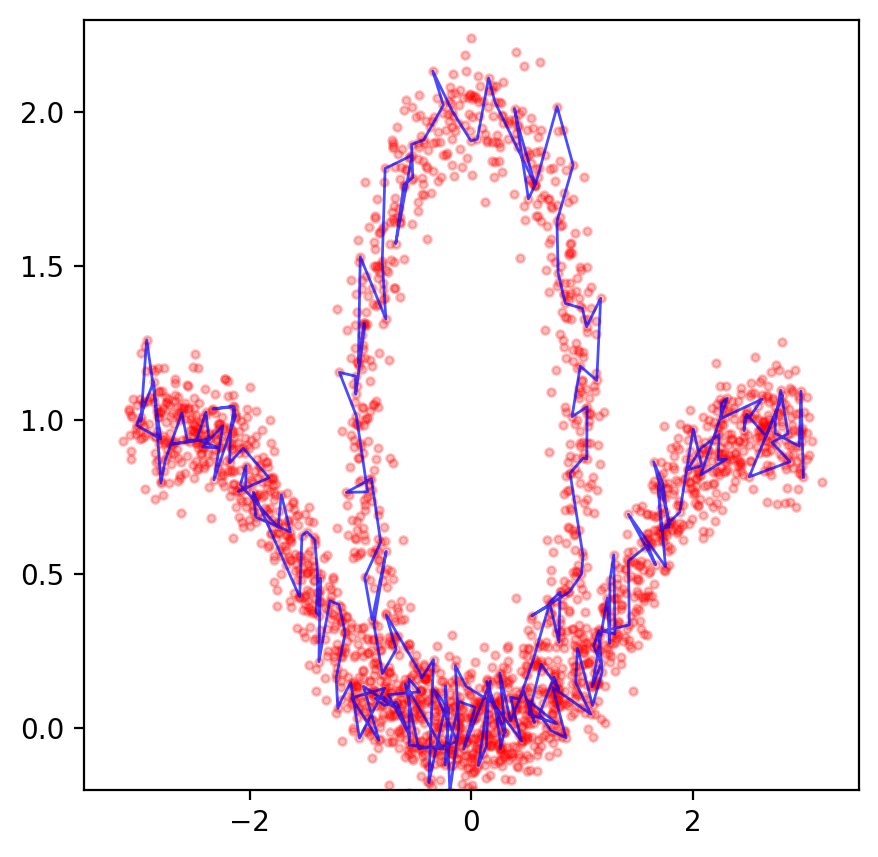
    }                                                                &
    \includegraphics[width=0.163\linewidth]{
      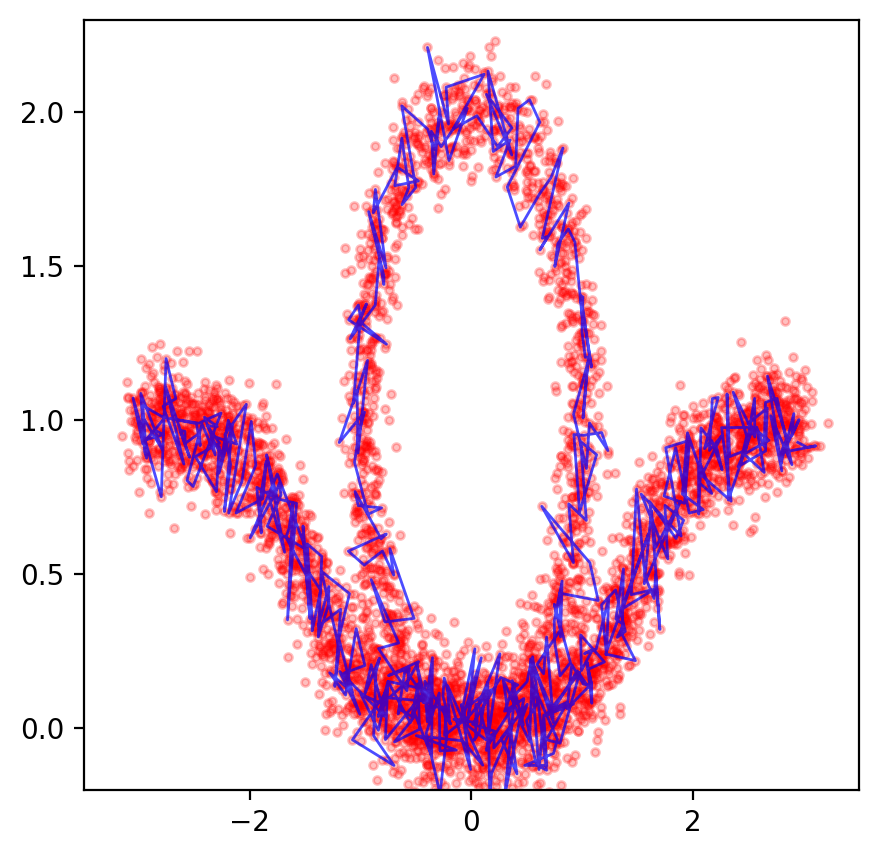
    }                                                                &
    \includegraphics[width=0.163\linewidth]{
      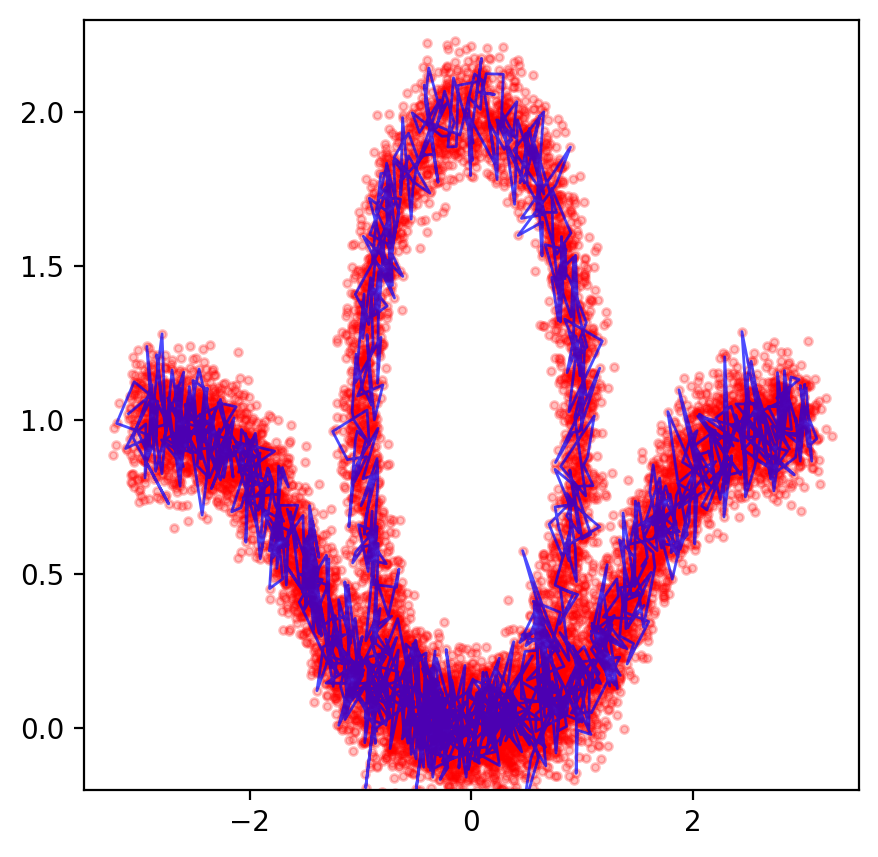
    }
    \\
    \raisebox{1.1cm}{\scriptsize \rotatebox[origin=c]{90}{ALI}}      &
    \includegraphics[width=0.163\linewidth]{
      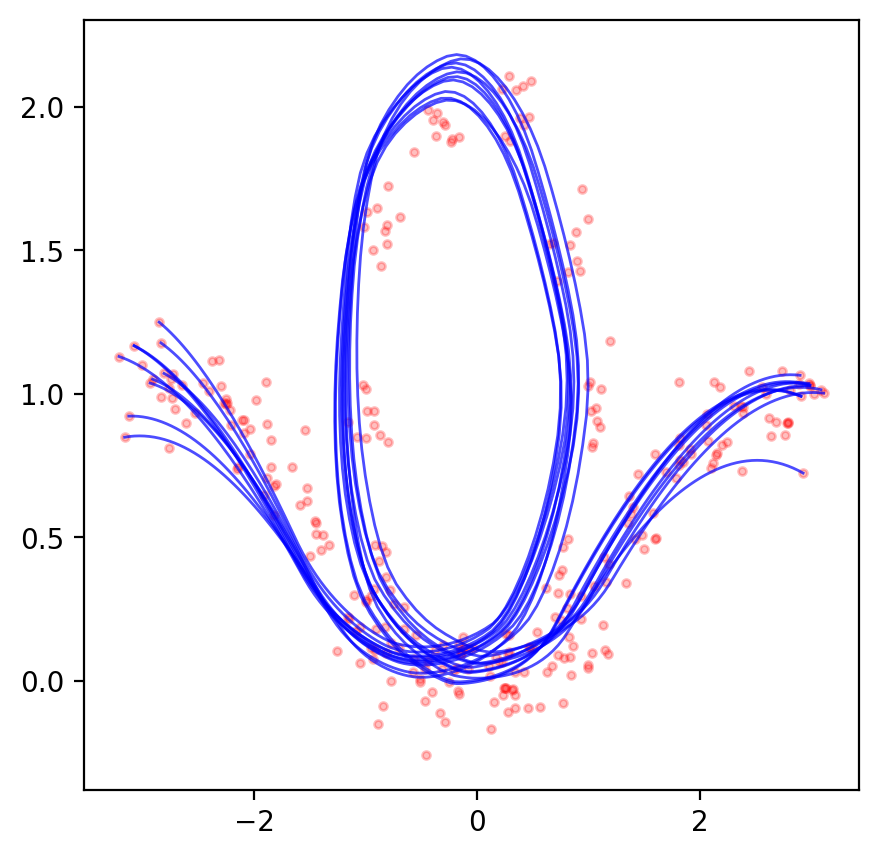
    }                                                                &
    \includegraphics[width=0.163\linewidth]{
      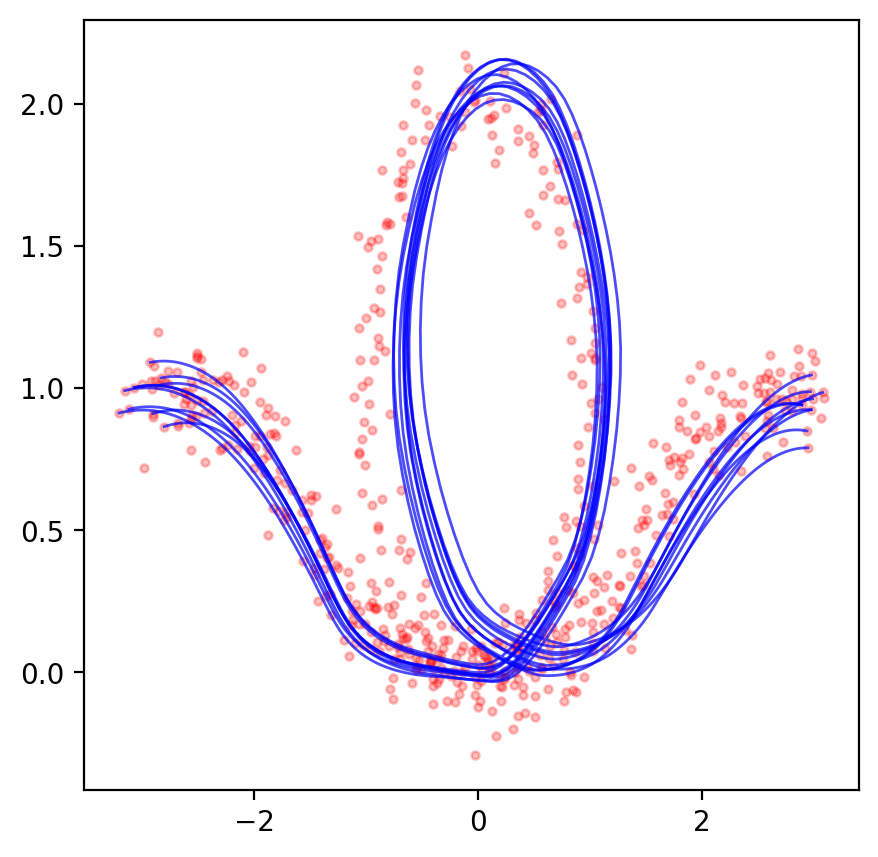
    }                                                                &
    \includegraphics[width=0.163\linewidth]{
      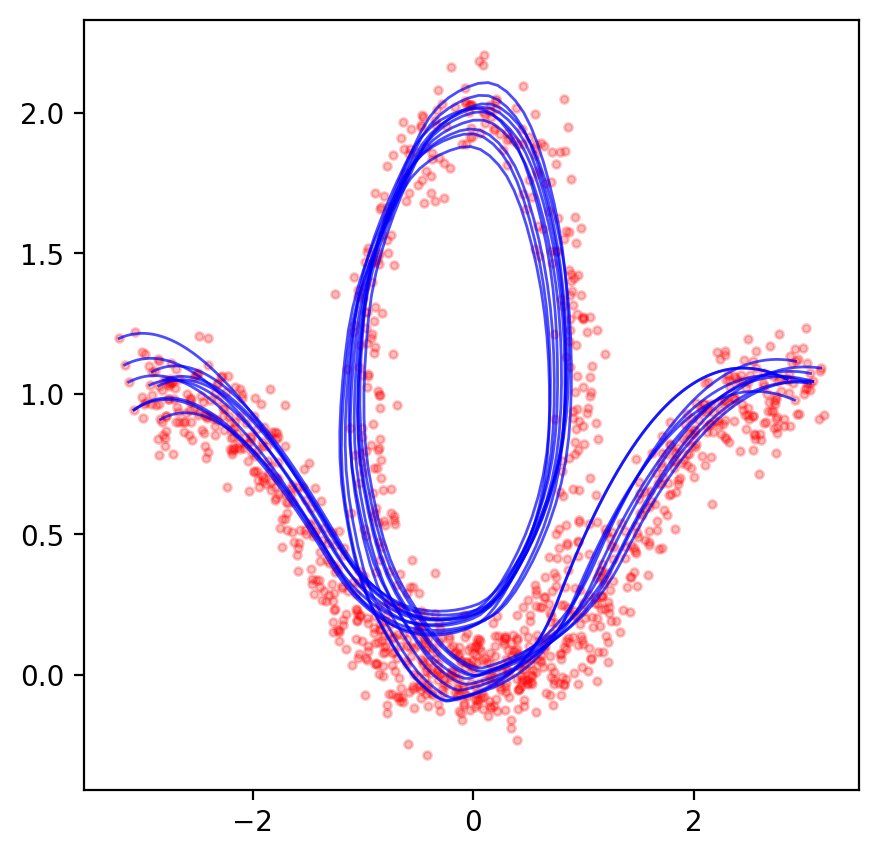
    }                                                                &
    \includegraphics[width=0.163\linewidth]{
      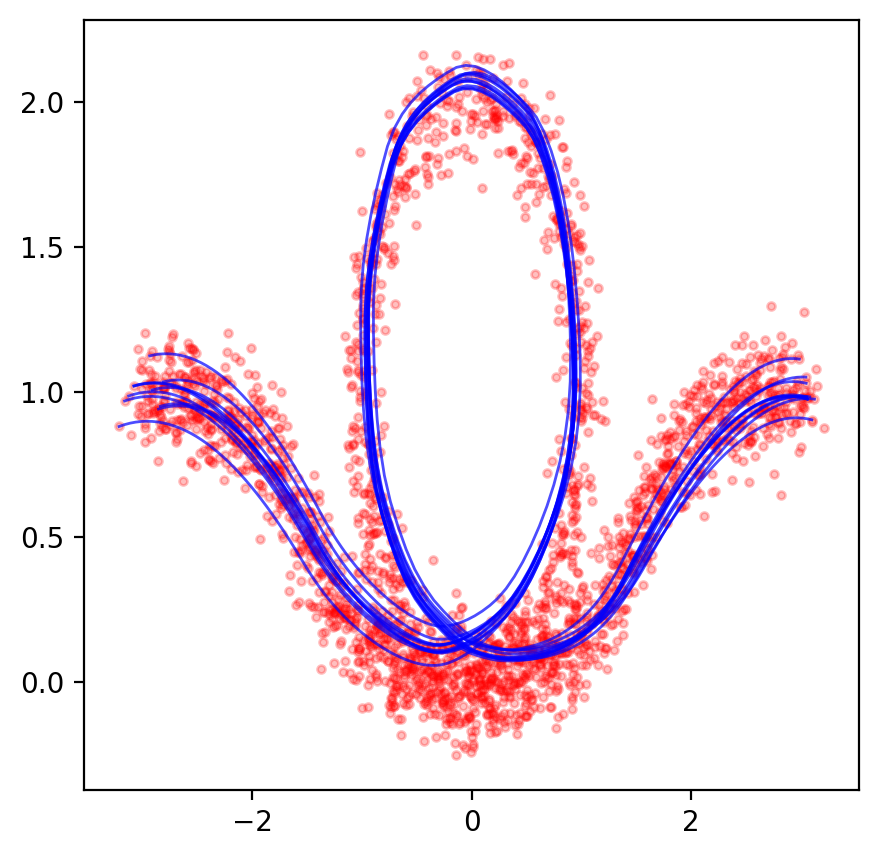
    }                                                                &
    \includegraphics[width=0.163\linewidth]{
      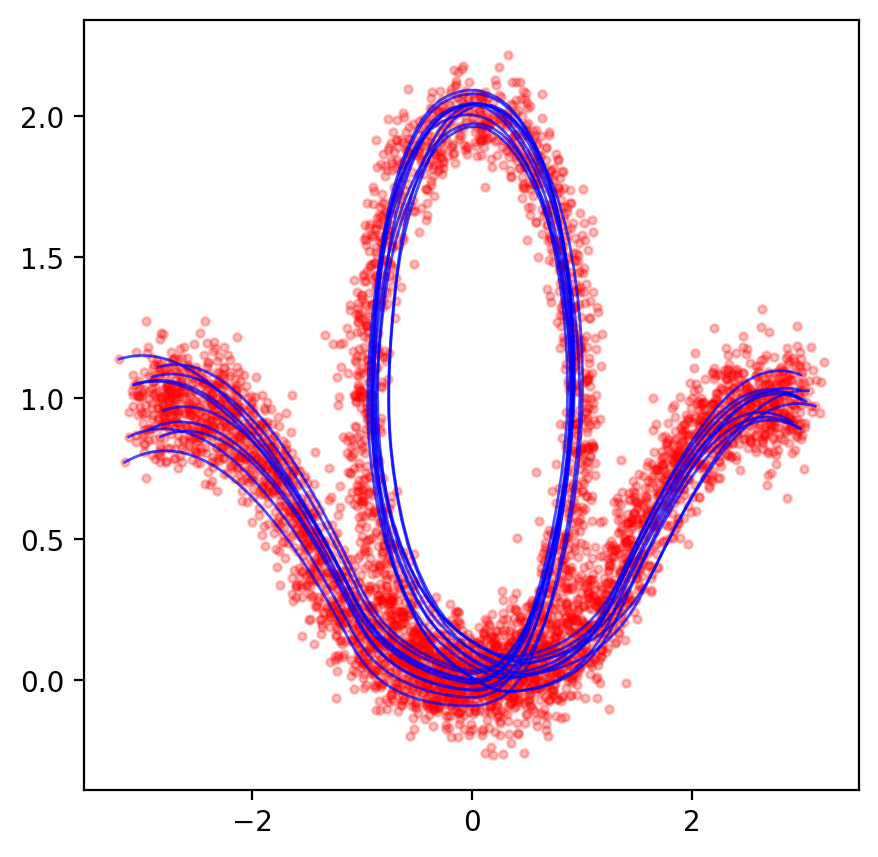
    }                                                                &
    \includegraphics[width=0.163\linewidth]{
      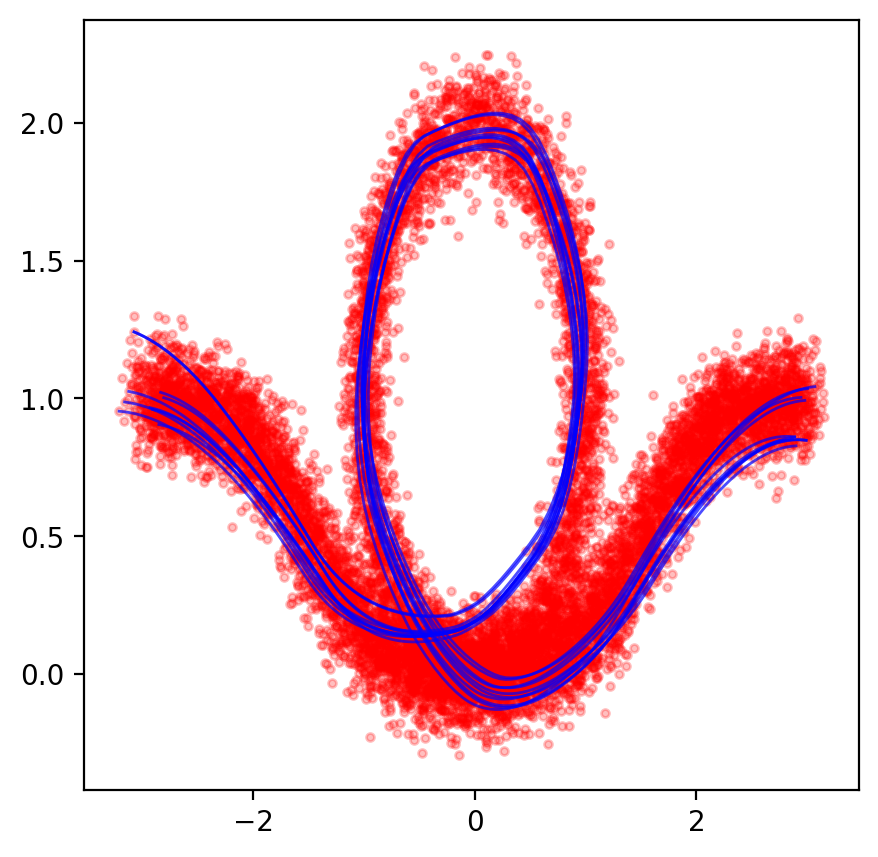
    }
    \\
  \end{tabular}

  \caption{
    Comparison between learnt time-dependent LAND-based interpolants
    used in OT-MFM, and the
    ALI interpolants for different number $K$ of marginal
    distributions on the knot dataset (\Cref{sec:knot_exp}).
  }
  \label{fig:interpolant-comparison}
\end{figure}

\paragraph{OT-MFM with time-dependent LAND metric}
We additionally ablate the use of LAND metirc in MFM
\citep{kapusniak2024metric} algorithm. \Cref{fig:interpolant-comparison} shows
how the learnt interpolants for MFM and ALI change as the number of marginal
distributions increases. When OT-MFM is equipped with a time-dependent LAND
metric, the interpolants are constrained to pass through the marginal samples
pointwise. As such, it is unsurprising that the resulting geopath interpolants
resemble the piecewise linear and cubic spline interpolants in
\Cref{fig:figure1}. As we see, the interpolants, trained using LAND MFM become
highly non-smooth, whereas ALI interpolants stay smooth regardless the number
of marginals.

As for the other methods in \Cref{fig:figure1}, we trained the vector field for
30,000 iterations with a learning rate of $10^{-4}$, and a two-layered MLP with
32 hidden units per layer and SELU activations.


\paragraph{OT-MFM with a time-dependent RBF metric}
Beyond the setting above, we also attempted to run MFM on the knot data with
the time-dependent RBF metric, as it is implemented in the publications'
associated code (see the single-cell trajectory inference section in
\cite{kapusniak2024metric} when the dimensions are 50 or 100D).

As outlined in \Cref{sec:knot_exp}, there are $K=1200$ marginal distributions
with ten samples from each distribution. Before we can train the MFM
interpolant (geopath) network, we therefore need to train $1199$ RBF networks
via gradient descent. 

We optimised the hyperparameters in the RBF training in order to jointly
optimise training time and metric loss, without getting NaN losses when
training the $1199$ networks. Each network is, by the construction of the
experiment and the time-dependent RBF implementation, fitted to $20$ samples.

With five clusters, hundreds of RBF nets had NaN losses independent of learning
rates. This happened due to numerical issues when computing the empirical
variance of the cluster-assigned points (line 76 in
\url{https://github.com/kkapusniak/metric-flow-matching/blob/main/mfm/geo_metrics/rbf.py}).
We found that two clusters did not result in NaN losses, and then used 50
training epochs per RBF net with the default learning rate, ensuring converging
loss curves. The training time—to fit the metrics, no training of the
interpolant or flow nets—was around 400 minutes, i.e., ~6.5 hours, on an NVIDIA
GeForce RTX 3080 GPU. Recall that this is 2D data.

\subsection{Cell tracking experiment}
\label{app:cell_tracking}

\begin{figure}
  \centering
  \begin{subfigure}[t]{0.52\linewidth}
    \includegraphics[width=1\linewidth]{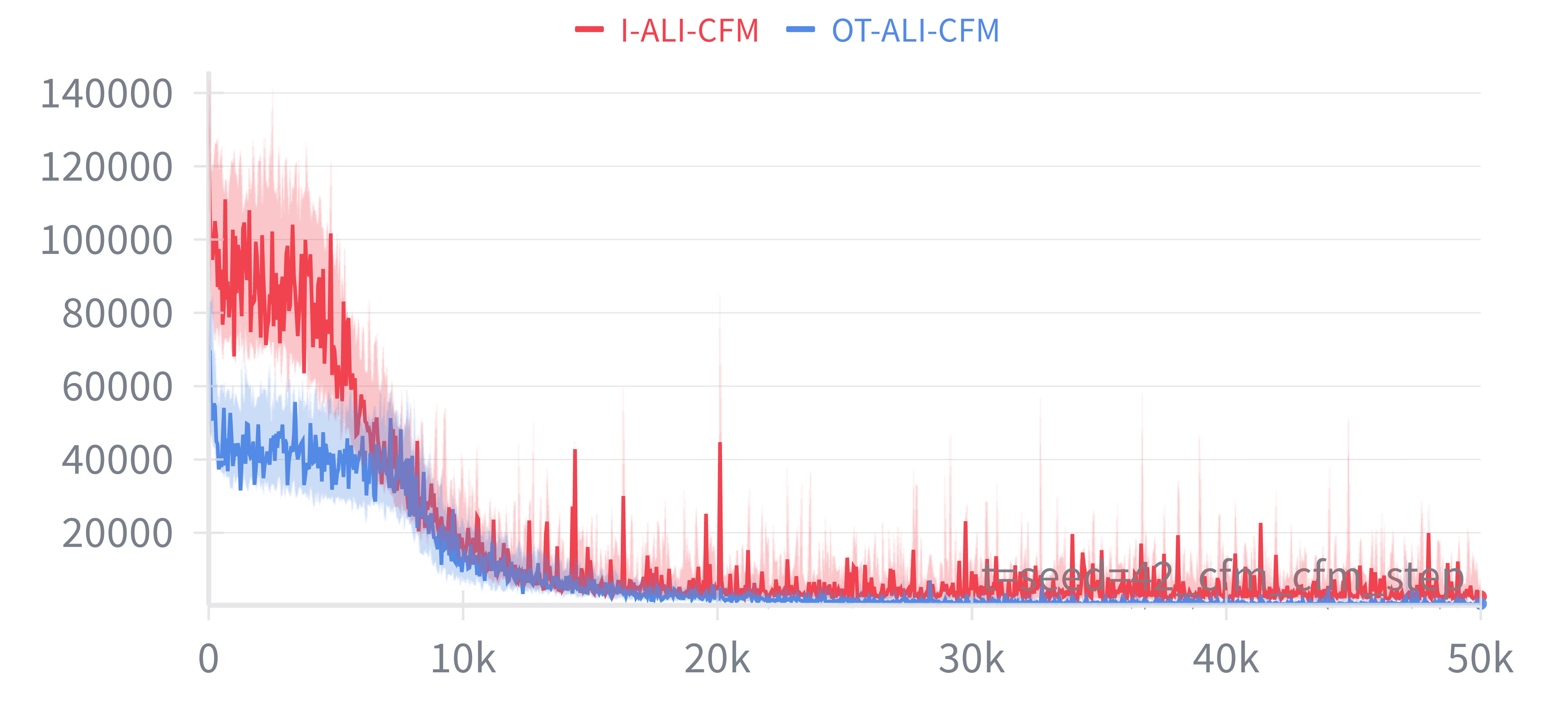}
    \caption{\textcolor{red}{I-ALI-CFM loss} vs.
    \textcolor{blue}{OT-ALI-CFM loss}}
  \end{subfigure}

  \begin{subfigure}[t]{0.32\linewidth}
    \includegraphics[width=1\linewidth]{
      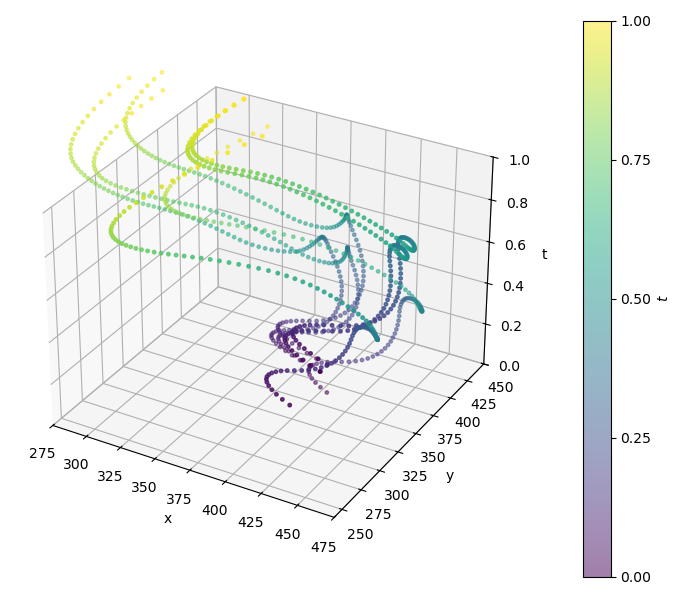
    }
    \caption{I-ALI}
  \end{subfigure}
  \begin{subfigure}[t]{0.32\linewidth}
    \includegraphics[width=1\linewidth]{cell_tracking_figures/3D_10_ali_interpolants.png}
    \caption{OT-ALI}
    \label{fig:ot_3d}
  \end{subfigure}
  \caption{
    Comparison of I-ALI-CFM and OT-ALI-CFM on the cell tracking data.
    As expected, the CFM losses in (a) show the increased variance in
    the CFM objective when using
    the independent coupling (red) compared to the OT coupling (blue).
    The effect of the increased variance is shown in (b) and (c), where the
    I-ALI trajectories overlap to a greater extent than the OT-ALI trajectories.
  }
  \label{fig:I_vsOT_cell_tracking}
\end{figure}
The data can be found on
\hyperlink{https://celltrackingchallenge.net/2d-datasets/}{this} website,
or explicitly via this link:
\url{https://data.celltrackingchallenge.net/training-datasets/PhC-C2DH-U373.zip}.
We use the frames available in folder \texttt{01} and the
corresponding segmentation masks from folder
\texttt{01\_ST/SEG}. We used the segmentation masks for the cell with
label four.

\paragraph{ALI hyperparamters.}
Both the interpolant and the discriminator are parameterised as 2-layered MLPs
with ELU activations and $256$ hidden units. We train ALI for $70,000$ epochs.
During training, we smooth the time-input to the generator by adding Gaussian
noise with standard deviation $10^{-3}$. Both networks are trained with
learning rates of $10^{-4}$ and separate Adam optimisers.

\paragraph{OT-MFM with a time-independent LAND metric.} For the cell
tracking experiments OT-MFM is trained
for $40,000$ epochs with
learning rate $10^{-4}$ for both interpolant and flow matching networks. The
weights of both networks are optimised with Adam. We use $\gamma = 0.4$ and
$0.9$ performing similarly, while leaving other hyperparameters of the LAND
metric default. Training for longer also does not yield better results with
these settings. Batch size is set to $128$. The neural networks have $128$
hidden units and $3$ hidden layers.

\paragraph{OT-MFM with a time-dependent LAND metric.}
For these experiments we trained both networks (same architecture as for other
OT-MFM experiments) for 75000 epochs and learning rate $10^{-4}$ (Adam); best
performing hyperparameters were $\gamma=0.2$ and other LAND hyperparameters set
default. While it would be possible to run OT every possible batch of
$(x_{t_{i}},x_{t_{i+1}})$ during training, we instead precompute OT maps
between the full successive pairs of marginals before training once and reuse
them, as the former option was too computationally expensive.
\paragraph{I-ALI-CFM vs. OT-ALI-CFM}
In \Cref{fig:I_vsOT_cell_tracking} we compare I-ALI-CFM and OT-ALI-CFM. As
expected, the CFM loss curves exhibit higher variance for I-ALI-CFM, while the
resulting I-ALI trajectories overlap to a greater extent than the OT-ALI
trajectories.

\subsection{scRNA-seq experiments}
\label{app:sc_experiments}
For single-cell experiments, we use the dataset available on
\url{https://data.mendeley.com/datasets/hhny5ff7yj/1},
the EB data used in \cite{tong2020trajectorynet} can be found on
\url{https://github.com/KrishnaswamyLab/TrajectoryNet/blob/master/data/eb_velocity_v5.npz}.
Following \cite{tong2023improving}, we additionally whiten the data
in the 5D experiments.

\paragraph{Hyperparameters for ALI-CFM.}
We parameterise learnable correction $f_\phi$ and $D_\gamma$ as two-layered
MLPs with $64$ (for the 5D experiments) or $1024$ (for the $50$ or 100D
experiments) hidden units in each layer with ELU activations. In accordance
with \cite{kapusniak2024metric}, we use a three-layered MLP with SELU
activations to model $u^\theta_t$. The number of hidden units in the layers of
$u^\theta_t$ is chosen to be either $64$ or $1024$, depending on the data
dimensions, as for $f_\phi$ and $D_\gamma$ above. We find that normalising the
data and adding small noise ($0.01 \cdot \epsilon$, $\epsilon \sim
\mathcal{N}(0, 1)$) to the time input of $G_\phi$ helps when training the GAN,
as it makes the interpolant smoother with respect to time input. At inference,
we push all the samples from marginal $i-1$ to the time associated with the
held-out marginal $t_{i}$, as in \cite{tong2023improving}.

\begin{figure}
  \centering
  \begin{subfigure}[t]{0.24\linewidth}
    \includegraphics[width=1\linewidth]{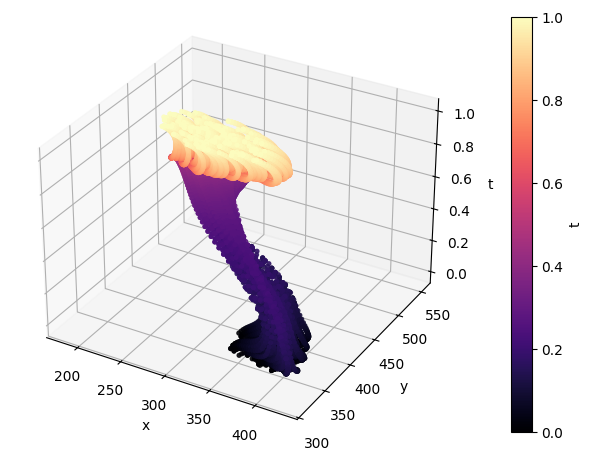}
    \caption{}
  \end{subfigure}
  \begin{subfigure}[t]{0.24\linewidth}
    \includegraphics[width=1\linewidth]{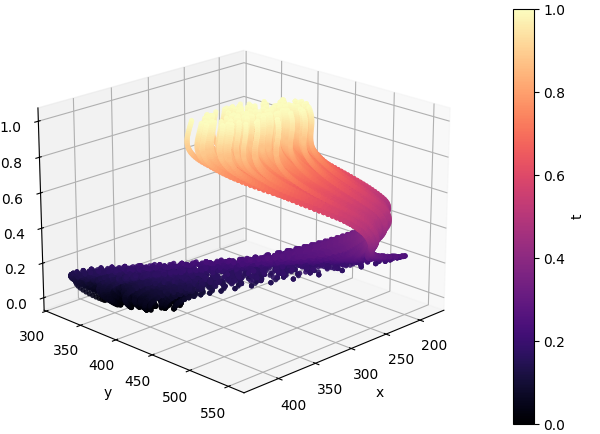}
    \caption{}
    \label{fig:mfm_3d_interpolants}
  \end{subfigure}
  \begin{subfigure}[t]{0.24\linewidth}
    \includegraphics[width=1\linewidth]{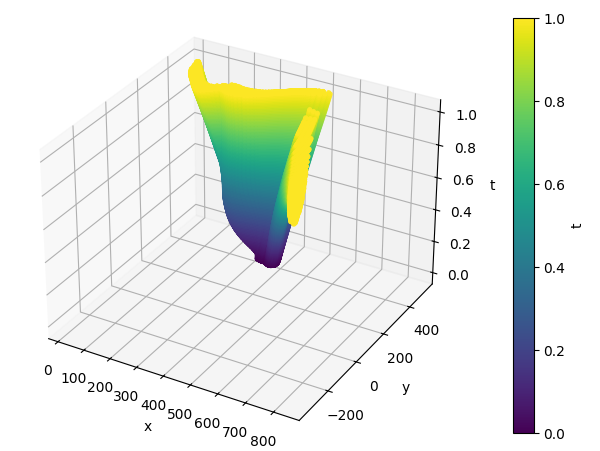}
    \caption{}
  \end{subfigure}
  \begin{subfigure}[t]{0.24\linewidth}
    \includegraphics[width=1\linewidth]{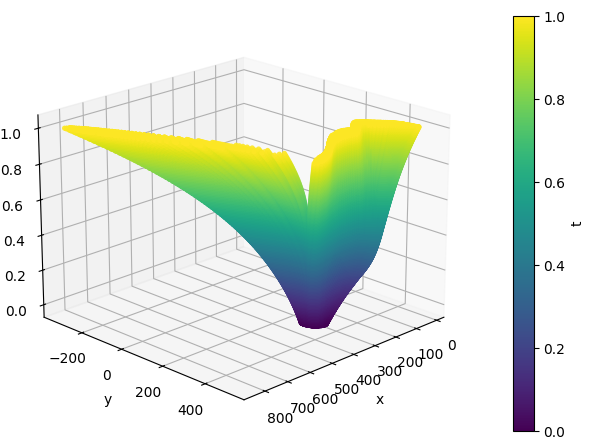}
    \caption{}
    \label{fig:otmfm_3d}
  \end{subfigure}
  \caption{Different 3D views of all time-independent OT-MFM
    interpolants are shown in \textbf{(a)} and \textbf{(b)}, while in
    \textbf{(c)} and \textbf{(d)} we show the 3D visualisations of the
    marginalised OT-MFM trajectories. In \textbf{(b)} we observe that
    the interpolants start to bifurcate around $t\approx 0.5$ and then
    abruptly change direction in order to satisfy the end-marginal
    constraint. We believe the bifurcation happens due to the guidance
    of the metric that was fitted on all training samples. Crucially,
    the bifurcation in the interpolation paths causes part of the
    vector field to diverge, which is seen in \textbf{(c)} and
    \textbf{(d)}. Note the extreme range of $x$-values in the 3D plots
    in \textbf{(c)} and \textbf{(d)} when mapping them to the scatter
  plots in \Cref{fig:cell_tracking_masks}.}
  \label{fig:mfm_cell_tracking}
\end{figure}

\begin{figure}
  \centering
  \begin{subfigure}[t]{0.34\linewidth}
    \includegraphics[width=1\linewidth]{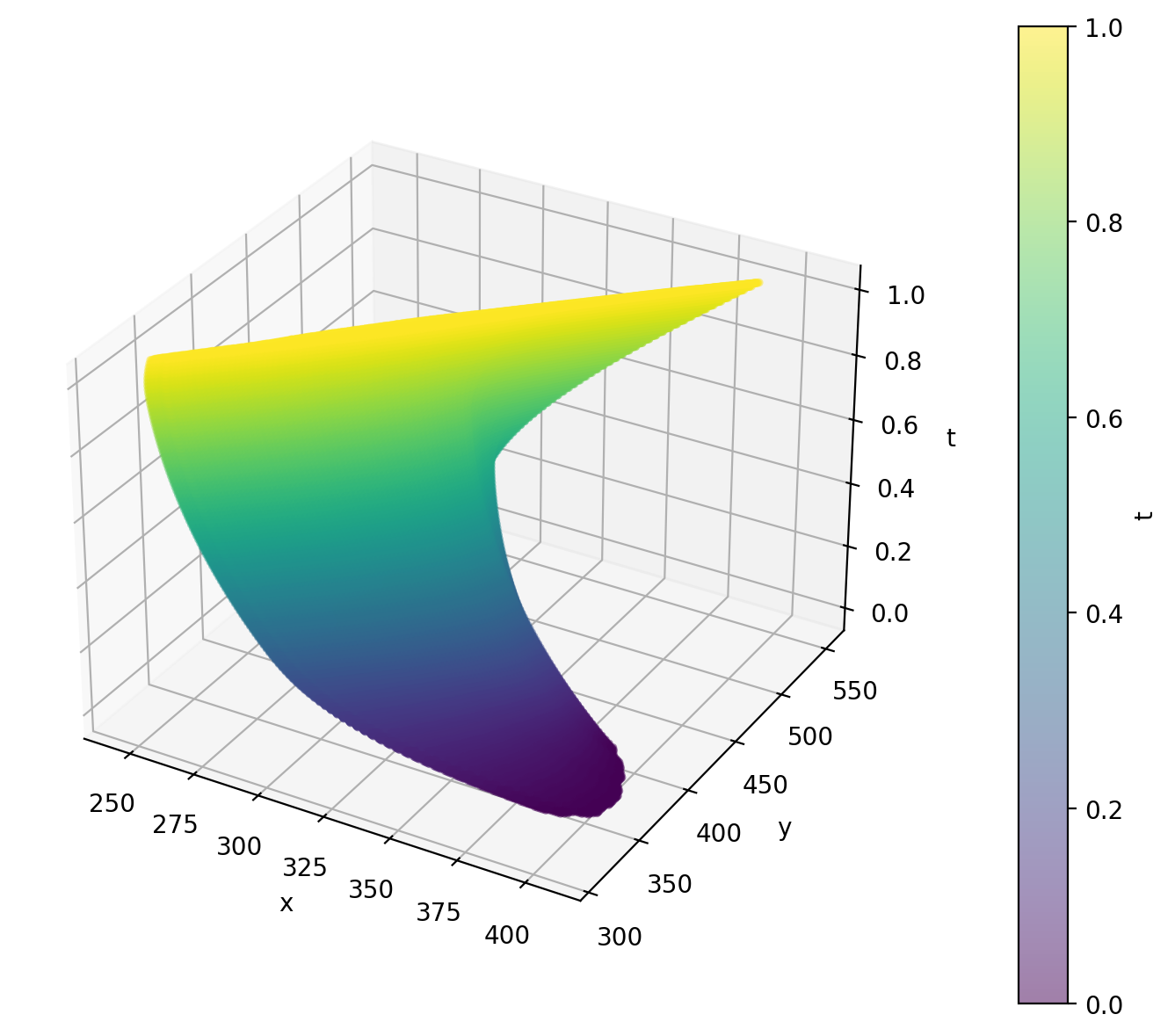}
  \end{subfigure}
  \begin{subfigure}[t]{0.75\linewidth}
    \includegraphics[width=1\linewidth]{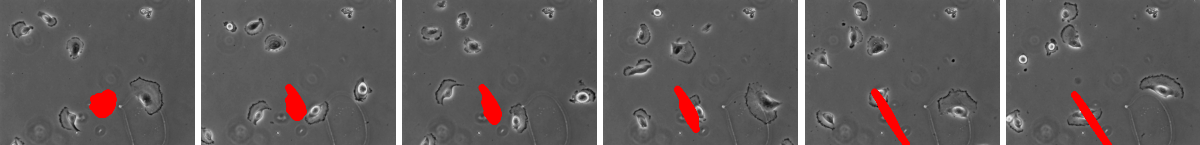}
  \end{subfigure}
  \caption{\textbf{Top:} The time-dependent OT-MFM marginalised
    vector field in 3D on the cell
    tracking data. \textbf{Bottom:} the trajectories overlayed on the
    a subset of the microscopy
  image frames (see \Cref{fig:cell_tracking_masks}).}
  \label{fig:time_dependent_otmfm_3d}
\end{figure}

For all of the single-cell datasets we pretrain the interpolant for $2,000$
steps and then train it using adversarial losses for $70,000$ steps. We
empirically found that the best results on single-cell data are obtained by
using R3GAN \cite{huanggan2024ganisdead} with the second-derivative
regulariser. We adjust the scale of the regulariser term in such a way that it
has the same value range as the generator loss. We use learning rate $10^{-3}$
for pretraining, $5\cdot10^{-5}$ for both discriminator and generator for
adversarial training and $10^{-3}$ for the OT-CFM training. We find that OT
coupling is important to achieve better results. We use multi-marginal OT
coupling in all our experiments. We also investigate the alternatives for the
$L_2$-norm. Following \citet{kapusniak2024metric}, we investigate substituting
the $L_2$-norm with the ``LAND'' metric in all of the previously discussed
regularisers. Given the dataset $\mathcal{D} = \{(x_{t_i}, t_i)\}_{i=1}^N $ and
$\epsilon > 0$ we let $(x_t, t) \mapsto g_{\rm LAND}(x_t, t) \equiv
G_{\epsilon}(x_t, t) = ({\rm diag}(h(x_t, t)) + \epsilon I)^{-1}$, where
\begin{equation}
  h_{\alpha}(x_t, t) = \sum_{(x_s, s) \in \mathcal{D}}
  (x_t - x_s)^2{\rm exp}\left( - \frac{\| x_t - x_s \|^2}{\gamma_1}
  \right){\rm exp}\left( - \frac{\| t -  s\|^2 }{\gamma_2} \right),
  \; 1 \leq \alpha \leq d,
\end{equation}
with $d$ being the dimension of $x$ and $\gamma_1$ and $\gamma_2$
being the kernel sizes.
Using ``LAND'' improved the results. We conduct our final experiments
on single-cell data using
the regulariser detailed in \Cref{eq:2nd_derivative} with the ``LAND'' metric.

\paragraph{Reproduced results.}
In order to fairly compare our method to the existing ones, we rerun all the
baselines on our hardware. The reported scores I/OT-MFM and I/OT-CFM are
obtained by running the code provided by \cite{kapusniak2024metric} without any
changes made to the code. For I-MMFM and OT-MMFM \citep{rohbeck2025modeling},
we implemented the cubic splines interpolants without the Gaussian probability
path, to align with the other considered methods. We only ran I/OT-MMFM on the
5D experiments, because this method poorly scales to the high dimensions.

\subsubsection{Ablation study for the different regularisation terms}
\label{app:ablation}
We conduct an ablation study on different dimensions of the CITE-seq
data, ablating
regularisers (\Cref{sec:regularisers}), regulariser coefficients, and
the effect of these on EMD,
computed on the left out time steps, and path energy, which
quantifies the straightness of the
trajectories (higher numbers, less straight paths) and is computed in
the following way:
\begin{equation}
  \text{path energy} = \mathbb{E}\left[\int_{0}^1\Vert
  u^\theta_{t}(x_t)\Vert^2 dt \;\Big|\; x_0 \sim q_0  \right].
\end{equation}
The results are shared in
\ThreeTabRef{tab:abl_pl}{tab:abl_2nd_50}{tab:abl_2nd_100}.

Interestingly, we see that increasing the regularising coefficient, combined
with the piecewise linear regulariser, makes the trajectories less smooth. We
argue that this is because the piecewise linearity forces interpolants to pass
through the marginal samples pointwise, akin to OT-CFM, MFMM and MFM with a
time-dependent metric, and so the trajectories become less straight.

Meanwhile, a stronger regularising weight combined with the other regularisers
in general straightens the interpolants (decreases the path energy). This is
expected as they directly penalise curvature.

\begin{table}[h]
  \centering
  \caption{
    EDM ($\downarrow$) for the learnt interpolant (Interpolant) and
    the marginalised vector
    field (CFM), as well as path energy ($\downarrow$) for the
    marginalised vector field (CFM path energy),
    computed on the left out time steps for Cite 50D and piecewise
    linear regulariser
  }
  \label{tab:abl_pl}
  \begin{tabular}{cc}
    Left out time step $t=1$                             & Left out
    time step $t=2$ \\
    \begin{tabular}{cccc}
      \toprule
      $\lambda$ & Interpolant & CFM   & CFM path energy \\
      \midrule
      0.01      & 46.96       & 47.29 & 300.01          \\
      0.05      & 41.11       & 41.07 & 697.19          \\
      0.10      & 40.76       & 41.19 & 1243.33         \\
      0.20      & 41.21       & 42.20 & 2979.58         \\
      0.50      & 41.82       & 40.93 & 5666.17         \\
      5         & 42.23       & 42.59 & 36264.89        \\
      10        & 41.38       & 43.81 & 87052.14        \\
      \bottomrule
    \end{tabular} &
    \begin{tabular}{ccc}

      \toprule
      Interpolant & CFM   & CFM path energy \\
      \midrule
      38.12       & 42.17 & 358.98          \\
      37.70       & 41.17 & 385.26          \\
      37.28       & 40.51 & 551.10          \\
      38.28       & 41.75 & 238.59          \\
      37.00       & 39.48 & 268.40          \\
      35.93       & 41.70 & 6297.56         \\
      35.57       & 43.06 & 8484.26         \\
      \bottomrule
    \end{tabular}                           \\
  \end{tabular}
\end{table}

\begin{table}[h]
  \centering
  \caption{
    EDM ($\downarrow$) for the learnt interpolant (Interpolant) and
    the marginalised vector field (CFM),
    as well as path energy ($\downarrow$) for the marginalised vector
    field (CFM path energy), computed
    on the left out time steps for Cite 50D and `the norm of the
    second derivative' regulariser
  }
  \label{tab:abl_2nd_50}
  \begin{tabular}{cc}
    Left out time step $t=1$                              & Left out
    time step $t=2$ \\
    \begin{tabular}{cccc}
      \toprule
      $\lambda$  & Interpolant & CFM   & CFM path energy \\
      \midrule
      1.00E{-06} & 41.50       & 42.17 & 12.30           \\
      5.00E{-07} & 41.48       & 42.14 & 12.00           \\
      1.00E{-07} & 41.44       & 42.06 & 12.56           \\
      5.00E{-08} & 41.38       & 42.23 & 13.91           \\
      1.00E{-08} & 40.75       & 41.59 & 26.00           \\
      5.00E{-09} & 41.39       & 42.24 & 41.98           \\
      1.00E{-09} & 42.89       & 43.81 & 56.39           \\
      \bottomrule
    \end{tabular} &
    \begin{tabular}{ccc}
      \toprule
      Interpolant & CFM   & CFM path energy \\
      \midrule
      39.63       & 45.66 & 7.58            \\
      39.55       & 45.51 & 7.55            \\
      39.35       & 45.23 & 7.68            \\
      39.31       & 44.99 & 8.10            \\
      39.06       & 41.61 & 13.21           \\
      39.04       & 40.14 & 18.51           \\
      41.24       & 40.04 & 39.65           \\
      \bottomrule
    \end{tabular}                            \\
  \end{tabular}

\end{table}

\begin{table}[h]
  \centering
  \caption{
    EDM ($\downarrow$) for the learnt interpolant (Interpolant) and
    the marginalised vector field (CFM),
    as well as path energy ($\downarrow$) for the marginalised vector
    field (CFM path energy),
    computed on the left out time steps for Cite 100D and `the norm
    of the second derivative' regulariser
  }
  \label{tab:abl_2nd_100}
  \begin{tabular}{cc}
    Left out time step $t=1$                              & Left out
    time step $t=2$ \\
    \begin{tabular}{cccc}
      \toprule
      $\lambda$  & Interpolant & CFM   & CFM path energy \\
      \midrule
      1.00E{-06} & 47.00       & 49.89 & 6.90            \\
      5.00E{-07} & 47.03       & 49.37 & 6.66            \\
      1.00E{-07} & 47.03       & 49.65 & 10.02           \\
      5.00E{-08} & 46.87       & 49.04 & 11.25           \\
      1.00E{-08} & 46.66       & 48.36 & 25.43           \\
      5.00E{-09} & 46.52       & 47.84 & 30.93           \\
      1.00E{-09} & 46.97       & 47.92 & 55.48           \\
      \bottomrule
    \end{tabular} &
    \begin{tabular}{ccc}
      \toprule
      Interpolant & CFM   & CFM path energy \\
      \midrule
      44.78       & 51.12 & 6.90            \\
      44.88       & 50.92 & 6.60            \\
      44.80       & 50.99 & 7.09            \\
      44.77       & 51.19 & 11.86           \\
      44.65       & 49.98 & 31.18           \\
      44.74       & 50.00 & 37.22           \\
      45.97       & 49.60 & 28.22           \\
      \bottomrule
    \end{tabular}
  \end{tabular}
\end{table}

\subsection{Spatial transcriptomics experiment}\label{app:st}
In \Cref{fig:inferred_BRCA}, we visualise some of the resulting tumour
coordinate inference results.

When marginalising the vector fields during inference of tumour coordinates at
the held-out time $t_i$, we push the samples from the observed $q_{t_{i-1}}$ to
$t=t_i$ using 101 integration steps and the \texttt{dopri5} ODE solver. For all
methods, the dataset is normalised to make the coordinates be in $[0, 1]$.
After training the CFMs, the marginalised CFM trajectories are denormalised and
compared with the unnormalised true data.

\paragraph{ALI hyperparameters.} We train $G_\phi$ and $D_\gamma$ for
$70,000$ epochs with the piecewise linear
reference in \eqref{eq:piecewise_reg}, and $\lambda=10$. We use a batch size of
$256$. Additionally, we pretrain $G_\phi$ for $1000$ epochs using a simple
$L_2$ regression objective between $G_\phi$ and samples from the observed
$q_{t_i}$. During training, we add a small Gaussian noise variable with
standard deviation parameter $0.001$ to the time input of the generator. The
learning rates in the adversarial training were $10^{-5}$ for both nets, and we
used the Adam optimiser with default hyperparameters.


\paragraph{CFM hyperparameters.} ALI-CFM is trained using a learning
rate of $10^{-3}$. Although we have tested
multiple learning rates for OT-CFM and OT-MMFM, the best performance is
attained using $10^{-3}$ for OT-CFM and $10^{-4}$. All nets are optimised using
the Adam optimiser.

\paragraph{OT-MFM hyperparameters.}

We train OT-MFM for $40,000$ epochs with learning rate $10^{-4}, 2 \cdot
10^{-4}$ and Adam for the interpolant and flow matching network respectively.
We note that we do not report results with the multi-marginal version of OT-MFM
using Eq. (20) of \cite{kapusniak2024metric}; in our preliminary tests, this
version exhibited worse results than the standard OT-MFM.

Batch size is set to $128$. We use the LAND metric as this is the recommended
one for low-dimensional data in \cite{kapusniak2024metric}. We set the
hyperparameters of the metric to $\gamma = 0.2$ for ST data and $0.4$ for cell
tracking data (other values perform worse, respectively); $\rho = 5 \cdot
10^{-4}$ for ST data seems to perform better than the default; the remaining
hyperparameters are set to MFM defaults. The neural networks have $128$ hidden
units and $3$ hidden layers.

\end{document}